\documentclass[10pt,journal,compsoc]{IEEEtran}
\usepackage{epsfig}
\usepackage{graphicx}
\usepackage{amsmath}
\usepackage{amssymb}
\usepackage{tabularx}
\usepackage{subcaption}
\usepackage{xcolor}
\usepackage{threeparttable}
\usepackage{algorithm}
\usepackage{algorithmic}
\usepackage{enumitem}
\usepackage{multirow}
\usepackage{lipsum}
\usepackage[hidelinks,pagebackref=false,breaklinks=true,colorlinks,bookmarks=false]{hyperref}
\ifCLASSOPTIONcompsoc
  \usepackage[nocompress]{cite}
\else
  \usepackage{cite}
\fi

\ifCLASSINFOpdf
\else
\fi

\usepackage{comment}

\usepackage{xspace}
\makeatletter
\DeclareRobustCommand\onedot{\futurelet\@let@token\@onedot}
\def\@onedot{\ifx\@let@token.\else.\null\fi\xspace}

\newcolumntype{P}[1]{>{\centering\arraybackslash}p{#1}}

\makeatletter
\renewcommand*{\p@section}{\S\,}
\renewcommand*{\p@subsection}{\S\,}
\makeatother

\newcommand{\erhao}[1]{\fontsize{11pt}{\baselineskip}\selectfont}

\usepackage{url}
\usepackage[utf8]{inputenc}
\usepackage{booktabs}
\usepackage[american]{babel}
\usepackage[T1]{fontenc}
\usepackage{balance}

\begin{document}
\title{Person Recognition in Aerial Surveillance: \\A Decade Survey}

\author{K.~Nguyen, 
        F.~Liu, 
        C.~Fookes, 
        S.~Sridharan, 
        X.~Liu, 
        A.~Ross
\IEEEcompsocitemizethanks{\IEEEcompsocthanksitem K. Nguyen, C. Fookes and S. Sridharan are with Signal Processing, AI and Vision Technologies (SAIVT) Research Program, Queensland University of Technology, Australia. \protect 
E-mail: \{k.nguyenthanh,c.fookes,s.sridharan\}@qut.edu.au}%
\IEEEcompsocitemizethanks{\IEEEcompsocthanksitem Feng Liu is with Department of Computer Science, Drexel University, United States. \protect 
E-mail: fl397@drexel.edu}
\IEEEcompsocitemizethanks{\IEEEcompsocthanksitem Xiaoming Liu and Arun Ross are with Department of Computer Science and Engineering, Michigan State University, United States. \protect 
E-mail: \{liuxm,rossarun\}@cse.msu.edu}
}

\markboth{IEEE T-BIOM}%
{Nguyen \MakeLowercase{\textit{et al.}}: Person Recognition in Aerial Surveillance}

\IEEEtitleabstractindextext{%
\begin{abstract}
The rapid emergence of airborne platforms and imaging sensors is enabling new forms of aerial surveillance due to their unprecedented advantages in scale, mobility, deployment, and covert observation capabilities. This paper provides a comprehensive overview of 150+ papers over the last 10 years of human-centric aerial surveillance tasks from a computer vision and machine learning perspective. It aims to provide readers with an in-depth systematic review and technical analysis of the current state of aerial surveillance tasks using drones, UAVs, and other airborne platforms. The object of interest is humans, where human subjects are to be detected, identified, and re-identified. More specifically, for each of these tasks, we first identify unique challenges in performing these tasks in an aerial setting compared to the popular ground-based setting and subsequently compile and analyze aerial datasets publicly available for each task. Most importantly, we delve deep into the approaches in the aerial surveillance literature with a focus on investigating how they presently address aerial challenges and techniques for improvement. We conclude the paper by discussing the gaps and open research questions to inform future research avenues.
\end{abstract}

\begin{IEEEkeywords}
Aerial surveillance, Eyes in the sky, Person Recognition at Altitude, Intelligence Surveillance and Reconnaissance
\end{IEEEkeywords}}

\maketitle

\IEEEdisplaynontitleabstractindextext

\IEEEpeerreviewmaketitle

\IEEEraisesectionheading{\section{Introduction}
\label{sec:introduction}}
Human-centric aerial surveillance is the task of employing airborne platforms, such as drones and aircraft, and imaging sensors mounted on them to monitor and gather data specifically related to human subjects. The primary objective of aerial surveillance is to develop computational models and techniques to detect, track, and identify, and monitor the behavior and activities of humans. This form of surveillance is critical for enhancing public safety, supporting law enforcement, and improving urban management by providing detailed and real-time insights into human movements and interactions.

Compared with standard ground-based surveillance, aerial surveillance offers distinct advantages in large-scale coverage, quick movement to a target, flexible deployment across day and night on any terrain, can be launched from a long distance, and establish covert presence \cite{EyesInTheSky}. However, developing efficient computer vision (CV) and machine learning (ML) models for person recognition in aerial surveillance also poses several unique challenges due to the novel perspectives and conditions of observation from the air. One significant challenge is the extreme viewing angles resulting from high-flying altitudes, where human subjects can appear in top-down views that are uncommon in traditional object detection and ground-based surveillance. Another notable challenge is the drastic variation in the size and scale of human subjects, who may appear extremely small due to the long distances between the aerial platforms and the subjects. Atmospheric turbulence further complicates the task by causing blurred or distorted visual data, which exacerbates the existing challenges for ML models. Addressing these challenges is the main focus of aerial surveillance research to advance the effectiveness and reliability of human-centric aerial surveillance systems.

Over the last decade, significant efforts have been made by the ML and CV communities, as well as the surveillance and defense sectors, to tackle these challenges. One of the key motivational milestones in aerial surveillance was the 2011 Gorgon Stare project, led by the Pentagon, which deployed an MQ-9 drone equipped with an advanced Multi-Spectral Targeting System called ARGUS to Iraq and Afghanistan for monitoring and tracking improvised explosive devices (IEDs). With as many as 368 individual cameras, Gorgon Stare could capture 1.8 billion pixels per frame, enough imaging power to spot an object six inches wide from an altitude of 7 km \cite{EyesInTheSky}. An entire city of size $10 \times 10$ km$^2$ could be continuously observed at a resolution sufficient to monitor any person or vehicle within the city. This eyes-in-the-sky system enabled persistent and mass surveillance of a wide area 24/7, even without the awareness of the city's citizens. 
\begin{figure*}
    \centering
    \includegraphics[width=1.8\columnwidth]{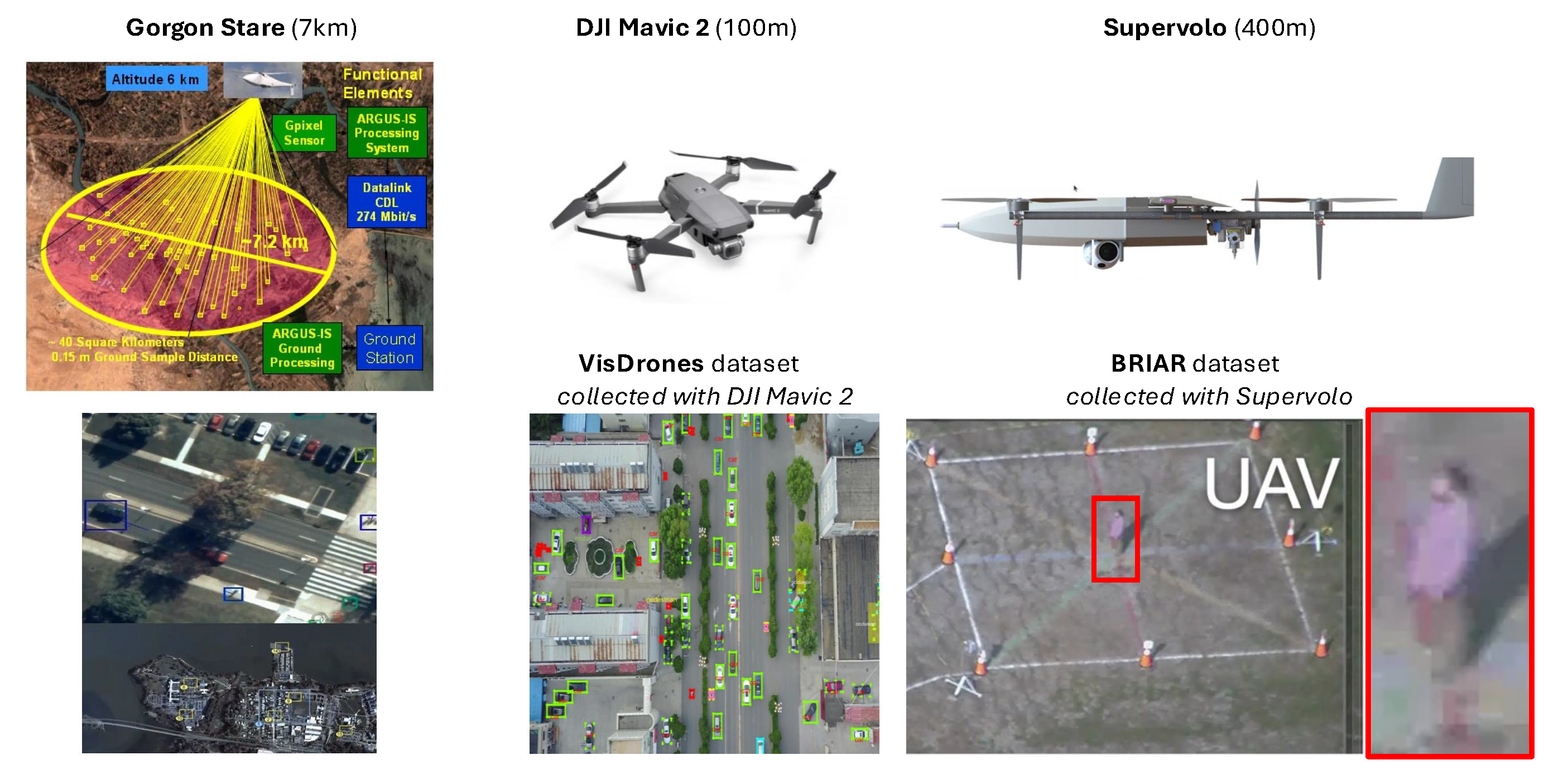}
    \caption{The rise of aerial surveillance. (1) The first column shows the eyes in the sky Gorgon Stare \cite{EyesInTheSky} deployed in Iraq, using an MQ-9 drone (costs \$$17M$) flying at an altitude of $7,000$ m, able to capture simultaneously an area of $10 \times 10$ km, with a $1.8$ GB sensor at a Ground Sampling Distance (GSD) of $15$ cm. (2) The second column is the off-the-shelf DJI Mavic 2 (costs \$$3K$) flying at an altitude of $100$ m. One example public dataset collected with DJI Mavic 2 was VisDrones \cite{VisDrones}. (3) The third column is Supervolo - a gasoline aircraft that can fly 8 hours at an altitude of $400$ m, able to capture an area of $1 \times 1$ km simultaneously, at a GSD of 20 cm. One example public dataset collected with Supervolo was BRIAR \cite{BRIARdataset}. }
    \label{fig:AerialSurveillance}
\end{figure*}
In parallel with military advancements, numerous academic initiatives have emerged. For example, five ``Vision meets drones'' challenges were organized in five consecutive years in major computer vision conferences (ECCV/ICCV 2018-2023). The challenges were based on a large-scale aerial dataset, VisDrones \cite{VisDrones} captured by the off-the-shelf DJI Mavic drones flying at an altitude of up to 100m, with more than 400 aerial videos formed by 265K frames and 2.6M bounding boxes or points of targets of frequent interests, such as pedestrians, cars, and bicycles. Another notable initiative is the BRIAR program \cite{BRIAR}, which was established in 2021 by the Intelligence Advanced Research Projects Activity (IARPA) to develop software algorithm-based systems capable of performing whole-body biometric identification at long ranges and from elevated platforms. The program released the BRIAR dataset in 2023 \cite{BRIARdataset} to facilitate aerial surveillance research under challenging scenarios, such as at long-range (\emph{e.g.} $300+$ meters), through atmospheric turbulence, with a rich set of airborne platforms (quadcopters and fixed wings) with altitudes up to 400 m and ranges up to 1000 m \cite{BRIARdataset}. These initiatives are illustrated in Fig.~\ref{fig:AerialSurveillance}.

\begin{figure}
    \centering
    \includegraphics[width=\columnwidth]{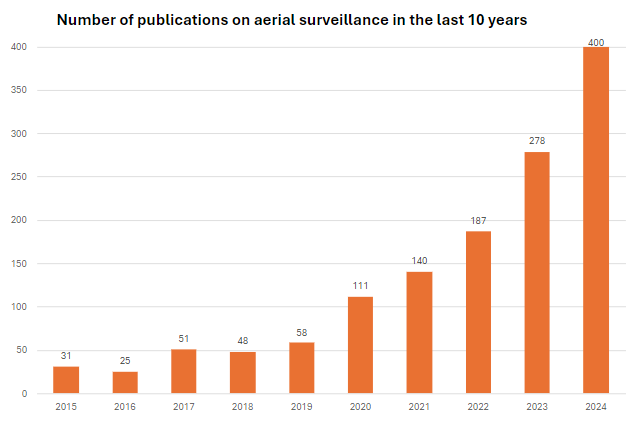}
    \caption{Increasing trend of publications in human-centric aerial surveillance over the last ten years. (\small{Data from Scopus advanced search: (TITLE (aerial OR uav OR drone) AND TITLE (detect* OR identi* OR recogni* OR re-id) AND TITLE (human OR person OR pedestrian OR object) AND NOT TITLE (vehicle)) AND PUBYEAR > 2014.}) }
    \label{fig:AerialSurveillancePapers}
\end{figure}

From the Gorgon Stare system to VisDrones and BRIAR, the uptake of aerial surveillance has broadened and advanced significantly, with academic research in this field experiencing a substantial boom. According to Scopus, there are more than 62,000 UAV/drone/aerial papers published 
and among them are 1600+ human-centric UAV/drone/aerial surveillance papers published 
in the last ten years, as illustrated in Fig.~\ref{fig:AerialSurveillancePapers}. 

Powered by recent breakthroughs in computer vision and deep learning, the implementation of the fundamental aerial surveillance tasks, such as detection, identification, and re-identification, on aerial data is emerging as an important and timely research area to be investigated. This paper dives deep into aerial surveillance tasks from computer vision, pattern recognition, and machine learning perspectives. While these tasks have been actively studied in both the generic computer vision and machine learning domains and the popular ground-based surveillance domain, the new challenges introduced by high-flying altitudes and the interdependencies with long ranges are daunting but targeted research to address these challenges is rapidly emerging. To systematically understand the state of human-centric aerial surveillance, we employ a structured framework to consistently investigate each task from four angles: 
\begin{itemize}
    \item (1) \underline{\textit{aerial surveillance challenges:}} identifies unique challenges for performing the task in the aerial domain, and investigates how well existing ground-based approaches shift to the aerial domain;
    \item (2) \underline{\textit{aerial surveillance datasets:}} compiles and analyzes the aerial datasets publicly available for the task;
    \item (3) \underline{\textit{approaches to solve aerial surveillance challenges:}} delves deep into the state-of-the-art approaches in the literature to study how they address the challenges unique to the aerial surveillance setting;
    \item (4) \underline{\textit{techniques to improve aerial surveillance tasks:}} further investigate other techniques in the literature to improve the performance of the aerial surveillance task.
\end{itemize}
The remainder of the paper is structured as follows. Section~\ref{sec:AdvantagesChallenges} identifies the advantages and challenges of aerial surveillance, then introduces a conceptual taxonomy to understand these challenges across ground-based and aerial surveillance. Sections~\ref{sec:HumanDetection}, ~\ref{sec:HumanIdentification},and ~\ref{sec:HumanReID}, investigate the state-of-the-art research of the three central aerial surveillance tasks: detection, identification, and re-identification. Section~\ref{sec:GapsOutlook} discusses challenges, open research questions, and future outlook for aerial surveillance. Section~\ref{sec:Conclusion} concludes the paper.


\section{Advantages, Challenges, and Taxonomy of Aerial Surveillance}
\label{sec:AdvantagesChallenges}
The evolution of aerial surveillance technology has revolutionized how we monitor and manage human activities from above. Leveraging advanced airborne platforms such as drones and UAVs, equipped with high-resolution imaging sensors, aerial surveillance offers unparalleled capabilities due to the high flying altitudes. This section delves into the distinct advantages of aerial surveillance while also highlighting the specific challenges that arise from recognising human subjects from aerial perspectives. Compared with standard ground-based surveillance, there are four distinct advantages to observation from the air:
\begin{itemize}
    \item \textit{Scale:} high-resolution imaging sensors on airborne platforms can cover vast areas from the air, providing detailed images at multiple resolutions. This reduces occlusion, allowing for more comprehensive monitoring compared to ground-based systems \cite{UAV-Human,TinyPersons}. 
    \item \textit{Mobility:} UAVs and drones can quickly reach and reposition over targets, offering dynamic and flexible surveillance. They can hover or circle over an area, ensuring continuous and optimal observation \cite{EyesInTheSky}. 
    \item \textit{Deployment:} airborne platforms can be deployed at any time, regardless of day or night, and in various terrains, including land and sea. Their ability to be launched from long distances makes them versatile and ready for immediate use in diverse situations \cite{EyesInTheSky,AVI}.
    \item \textit{Covert observation:} airborne platforms can adjust the flying altitudes to either remain hidden from view or make their presence known for direct intervention. This flexibility allows for both discreet monitoring and overt actions when necessary \cite{EyesInTheSky}.
\end{itemize}
However, aerial surveillance introduces a plethora of challenges that must be addressed. It not only inherits the existing challenges of unconstrained and outdoor surveillance but also presents additional unique challenges due to its distinct setting. From an analysis perspective, there are nine key challenges to contend with:
\begin{itemize}
    \item \textit{Low resolutions:} subjects may appear extremely small due to the high flying altitude, making it difficult to identify and track them accurately  \cite{TinyPersons,VisDrones}.
    \item \textit{Multiple scales:} multiple instances of the same class, \emph{e.g.} person, can appear drastically different in sizes and scales within the same scene \cite{SAMR,ScaleInvarianceAerial}.
    \item \textit{Extreme views:} objects can appear in overhead views, \emph{i.e.} top views and angle views, which are uncommon in generic object detection and require specialized algorithms 
    \cite{GLSA,AGVPReID2025}.
    \item \textit{Moving cameras:} view of objects may change continuously due to the movement of the camera mounted on the airborne platform \cite{UAV-Human}. This adds challenges of motion blur and camera stabilization \cite{FarSight,EyesInTheSky}. 
    \item \textit{Non-uniform distribution:} subjects are often distributed non-uniformly, either clustered with high density in busy areas like urban villages \cite{ClusterDet,ClusterNet}; or scattered with low density over wide areas, such as in search and rescue operations \cite{NDFT}. 
    \item \textit{Illumination:} variations in local illumination and lighting conditions across a wide area can affect image quality, requiring robust handling of both strong and low light conditions \cite{VisDrones}.
    \item \textit{Noise Artifacts:} weather conditions such as clouds, fog, haze, and rain can obstruct the scene, introducing noise artifacts that complicate data analysis \cite{DomainLabels,NDFT}.  
    \item \textit{Occlusion and Self-Occlusion:} subjects may be partially occluded by other objects or self-occluded due to the angle of view, making detection and identification more challenging \cite{VisDrones}.
    \item \textit{Atmospheric turbulence:} imaging at long ranges can suffer from atmospheric turbulence, leading to unavoidable degradation in image quality \cite{FarSight, BRIARdataset}.
\end{itemize}
These challenges highlight the complexity of developing effective aerial surveillance systems. Addressing these issues requires innovative approaches in computer vision and machine learning to enhance the accuracy and reliability of human-centric aerial surveillance.
\begin{figure}[H]
    \centering
    \includegraphics[width=0.9\columnwidth]{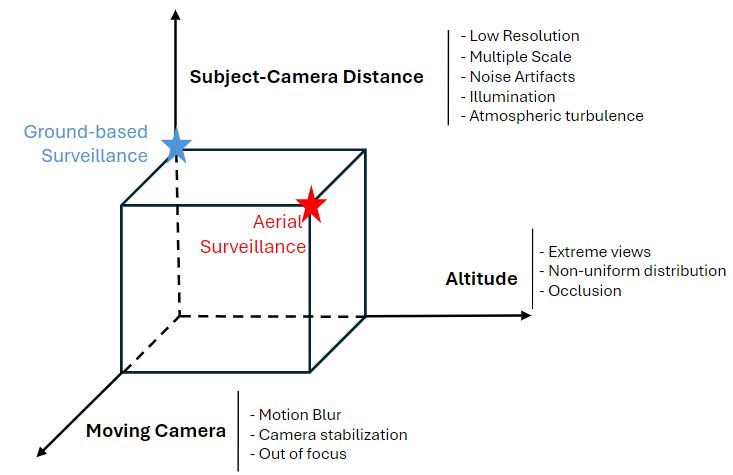}
    \caption{Three challenge dimensions of surveillance—subject-camera distance, altitude, and camera motion - highlighting the contrast between aerial and ground-based systems.}
    \label{fig:Aerial_Challenges_Dimensions}
\end{figure}

\begin{figure*}
    \centering
    \includegraphics[width=1.8\columnwidth]{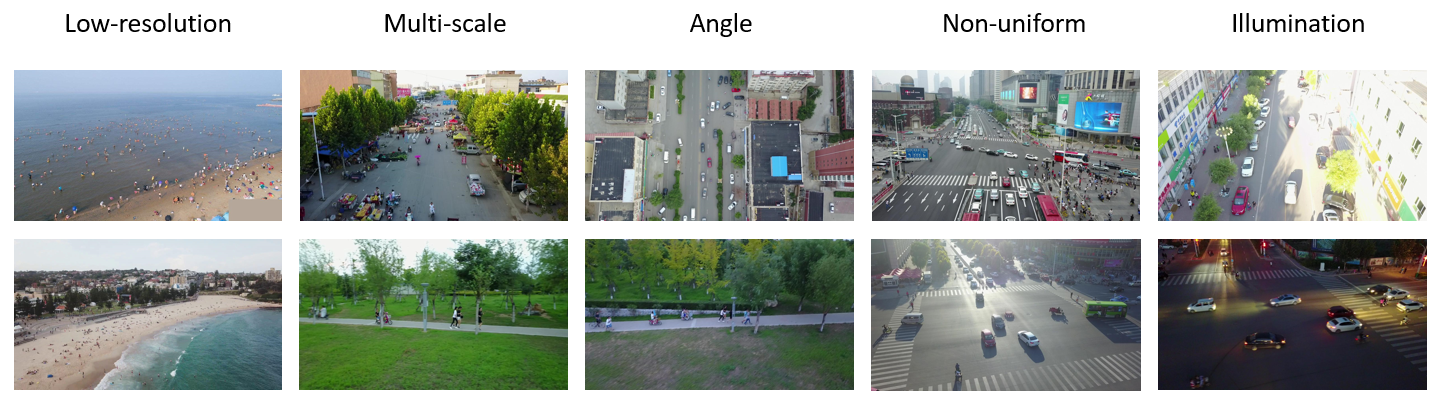}
    \caption{Challenges for aerial object detection: (i) low resolution, (ii) a wide range of scales, (iii) arbitrary viewing angles, (iv) non-uniformly distributed, (v) illumination. Images from the TinyPersons \cite{TinyPersons} and VisDrone \cite{VisDrones} datasets.}
    \label{fig:AerialChallenges}
\end{figure*}

Building on these observations, we propose a conceptual taxonomy to better understand how these challenges interrelate and compound in aerial surveillance settings. The taxonomy is structured around three interrelated dimensions: subject-camera distance, camera motion, and altitude. These factors collectively shape the complexity of aerial observation and are often intertwined in their effects. For instance, increased altitude not only reduces image resolution but also heightens the risk of occlusion and atmospheric distortion, which in turn degrade detection and identification accuracy. Similarly, camera motion - common in airborne platforms - introduces motion blur and instability, complicating both detection and recognition tasks. Unlike ground-based surveillance, which typically operates along a stable vertical axis, aerial systems are subject to dynamic conditions that amplify existing challenges and introduce new ones. Figure~\ref{fig:Aerial_Challenges_Dimensions} illustrates how these dimensions interact, underscoring the need for resilient and adaptive solutions tailored to aerial environments.


\section{Aerial Human Detection}
\label{sec:HumanDetection}

Aerial human detection is a pivotal component of aerial surveillance, serving as the foundation for subsequent tasks. This task addresses two fundamental questions: \textit{Are there humans present in the scene, and if so, where are they located?} The inherent complexity of detecting humans arises from the non-rigid nature of the human body and the variability in appearance due to different outfits, body shapes, and dynamic movements. These challenges are further exacerbated in aerial surveillance due to the high altitudes and extensive ranges involved. As generic human detection has been well established, it is not strange that most efforts in aerial human detection are spent on the extreme challenges of the aerial setting such as tiny resolution, extreme top view angle, and non-uniform distribution with humans significantly different in sizes in the same image, and extreme illumination. Effective aerial human detection systems must be capable of accurately and reliably identifying humans under these demanding conditions, ensuring robust performance across diverse scenarios.


\vspace{-6px}
\subsection{Challenges for aerial human detection}
\label{sec:AerialHumanDetChallenges}
\begin{figure}
    \centering    \includegraphics[width=0.8\columnwidth]{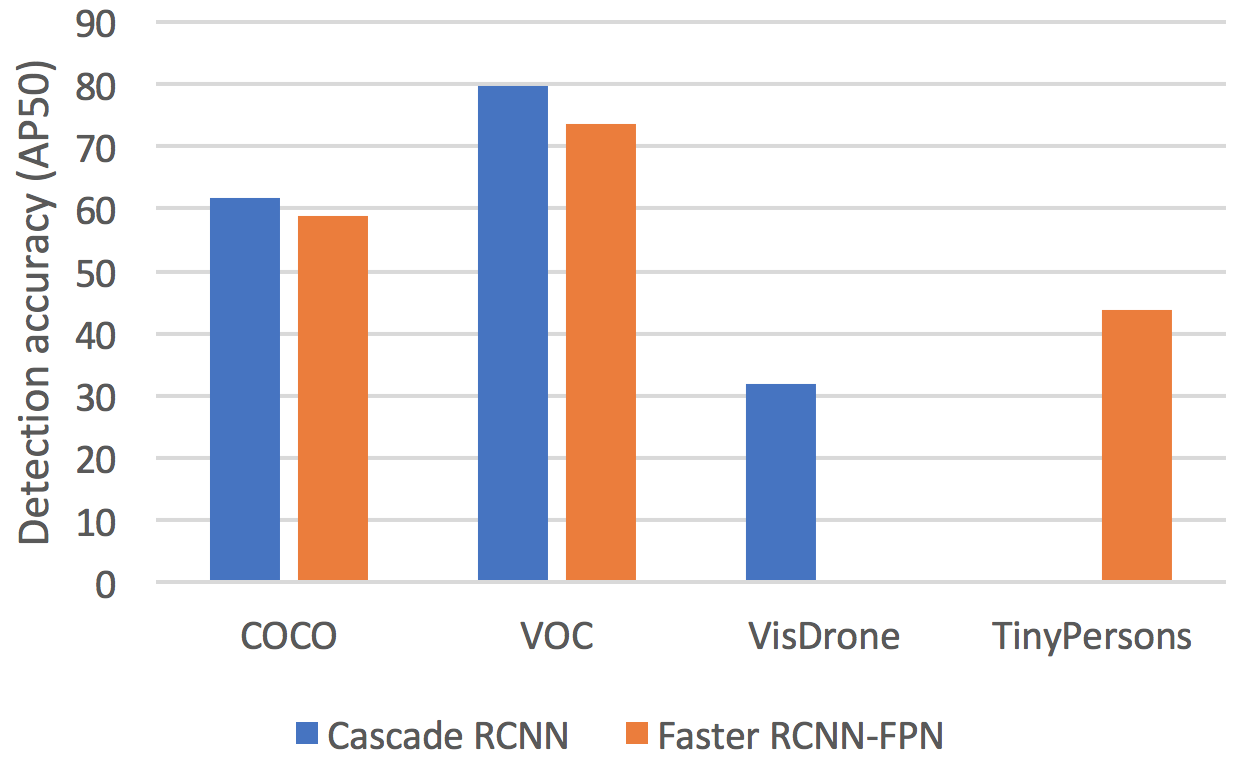}
    \caption{State-of-the-art generic detectors drop the performance when shifting to the aerial data. The figure compares the detection accuracy (AP50) of Cascade R-CNN \cite{CascadeRCNN} and Faster R-CNN \cite{FasterRCNN} on two ground-based (COCO and VOC) and two aerial (VisDrone and TinyPersons) datasets.}
    \label{fig:AerialDetectionPerformance}
\end{figure}

Despite the great success of the generic object detection methods trained on ground-to-ground images, a huge performance drop is observed when they are directly applied to images captured by UAVs \cite{VisDroneDET2020,TinyPersonsChallenge}. Examples of the performance drop of state-of-the-art detectors are illustrated in Fig.~\ref{fig:AerialDetectionPerformance}. The Cascade R-CNN \cite{CascadeRCNN} drops the performance by 50\% in the aerial VisDrone \cite{VisDrones} dataset compared to the ground-based COCO and Pascal VOC datasets \cite{VisDroneDET2020}. The Faster R-CNN \cite{FasterRCNN} also drops the performance by 30\% in the aerial TinyPersons \cite{TinyPersons} dataset compared to the ground-based COCO and Pascal VOC datasets \cite{TinyPersonsChallenge}. The performance drop is owing to the domain shift when compared to ground-based data caused by the high flying altitude and the camera characteristics. 

As human detection usually requires fewer details of humans than other surveillance tasks, it can be performed at extremely high altitudes and long ranges. In extreme cases, a human may only occupy several pixels in their absolute size\footnote{the absolute size of an object is defined as the square root of the object’s bounding box area.} due to the high-flying altitudes and wide-angle lens, as illustrated in Fig.~\ref{fig:AerialChallenges}. The high spatial resolution and large coverage of the aerial cameras can cause subjects to appear in significantly different sizes, different distribution uniformity, and potentially illumination non-uniformity in the same image, as illustrated in Fig.~\ref{fig:AerialChallenges}. The high altitudes and long ranges amplify all seven challenges discussed in Section~\ref{sec:AdvantagesChallenges} for aerial human detection.

\vspace{-6px}
\subsection{Datasets for aerial human detection}
The number of aerial datasets has quickly increased in the last few years, partially due to the affordable availability of off-the-shelf drones such as DJI. There are two categories of aerial human detection datasets: (1) only human class, and (2) human as one among other classes. Most datasets in Category 1 are dedicated to search and rescue. We compile a list of public datasets and their statistics in Table~\ref{tab:HumanDetectionDatasets}. Among a wide range of public datasets, two notable large-scale ones are VisDrone \cite{VisDrones} and TinyPersons \cite{TinyPersons}. 
\begin{itemize}
  \item \textit{VisDrone} \cite{VisDrones}: the VisDrone team has compiled a dedicated large-scale drone benchmark and organized five challenges in object detection in ECCV/ICCV from 2018 to 2023. It consists of 400 video clips formed by 265K frames and 10K static images with 2.6M bounding boxes, captured by various drones flying at 30 to 100 m, covering a wide range of aspects including location (14 cities), environment (urban and country), objects (pedestrian, vehicles, bicycles, \emph{etc.}), and density (sparse and crowded scenes). 
  \item \textit{TinyPersons} \cite{TinyPersons}: this dataset is behind two ``Tiny Object Detection'' challenges in ICCV 2019 and ECCV 2020. The unique characteristic of this dataset is the tiny resolution of humans. A majority of human instances appear as small as $[20,32]$ pixels and as tiny as $[2,20]$ pixels. 
  The tiny human sizes are due to the high flying altitudes of drones over crowded beaches with wide-angle camera lenses.
\end{itemize}
There is a wide range of other datasets with specific characteristics targeting various applications. 
\begin{itemize}
    \item Context: there is a diverse range of contexts, from urban villages \cite{VisDrones,HERIDAL,UAV123}, rural \cite{VisDrones,UAV-Human}, university campuses \cite{StanfordDrones}, Internet \cite{TinyPersons}, forest \cite{BIRDSAI}, to agricultural areas \cite{AgriDrone}.
    \item Flying altitudes: there is a diverse range of flying altitudes, from a very close distance of less than 10m \cite{PDESTRE}, a close range of 10-50m \cite{UAV123,StanfordDrones}, to a middle range of 50-120m \cite{AgriDrone,UAV-Human}, and to above the recreational limit >120m \cite{TinyPersons,Seadronessee}. 
\end{itemize}

\noindent\textbf{Multimodal:} many datasets also provide multi-modal data from other on-board sensors such as GPS, time, altitude, IMU, velocity, and weather conditions. The AU-AIR \cite{AUAIR} labeled each frame with time, GPS, IMU, altitude, and velocities of the UAV. \cite{NDFT} annotated and released altitude, viewing angle, and weather for the VisDrone dataset. These auxiliary details can be used to improve detection \cite{NDFT}.

\vspace{3px}
\noindent\textbf{Task-specific:} many aerial human detection datasets are aimed for specific tasks. The HERIDAL \cite{HERIDAL,HERIDAL2} and SARD \cite{SARD} datasets were collected to support the search and rescue mission with drones. The AgriDrone \cite{AgriDrone} dataset was collected in an agricultural context for rural applications.

\vspace{3px}
\noindent\textbf{Beyond visible:} sensors from other spectrums have been employed to complement visible cameras. Infrared sensors complement visible cameras in such adverse conditions as nighttime or low light. Both the BIRDAI \cite{BIRDSAI} dataset and UAV-Human \cite{UAV-Human} dataset provide thermal-infrared images of humans for detection. Depth sensors complement visible cameras in 3D object details and can be found in the recent UAV-Human \cite{UAV-Human} dataset.

\vspace{3px}
\noindent\textbf{Competitions:} the task of aerial human detection has attracted great attention from the computer vision community. Multiple challenges have been organized in top-tier conferences. The challenges ``Vision meets drones'' \cite{VisDroneDET2020}, including detection, tracking, and crowd counting, have been organized in five consecutive years in ECCV/ICCV from 2018 to 2023 using the VisDrone dataset. The TinyPerson challenges \cite{TinyPersonsChallenge}, organized in ICCV 2019 and ECCV 2020, focus on persons from a very long distance with a wide view using the TinyPerson dataset. The Marine Computer Vision challenges \cite{MaCVi, MaCVi2, MaCVi3}, organized in WACV 2023 to 2025, focus on detecting swimmers from drones in the marine context of search and rescue using the SeaDronesSee Object Detection dataset \cite{Seadronessee}.

\begin{table}[t]
\small
\caption{Public datasets for aerial human detection. `Alt.', `\#Con', `\#Img', and `\#Box' respectively represent the altitudes in meters, the context where data was collected, the number of images (train+test), and bounding boxes.}
\label{tab:HumanDetectionDatasets}
\centering
\resizebox{\columnwidth}{!}{
\begin{tabular}{l l l c c r r r r}\toprule
{Dataset} & {Year} & {Alt.(m)} & {Con.} & {\#Img.} & {\#Box.}\\
\midrule
\midrule
  StanfordDrone~\cite{StanfordDrones} & 2016  & 80 & Campus &- &11K\\
  HERIDAL~\cite{HERIDAL}            & 2019    & 40-65  & Natural &68K &-\\
  TinyPerson~\cite{TinyPersons}     & 2020    & -  & Internet & 1.6K & 72K  \\
  AU-AIR~\cite{AUAIR}               & 2020    & 5-30  & Real &32K &132K\\
  AgriDrone~\cite{AgriDrone}    & 2020  & -  & Agricultural &1.6M &-\\
  SARD~\cite{SARD}                  &  2021   & 5-50  & Natural &4K &6.6K\\
  VisDrone~\cite{VisDroneDET2021}   & 2018    & 30-100  & Urban   & 10K & -  \\
  BIRDSAI~\cite{BIRDSAI}        & 2020  & 60-120 & Forest &162K & 84K\\
  UAV-Human~\cite{UAV-Human}    & 2021  & 2-8 & Various  & 64K & 41K\\
  Seadronessee~\cite{Seadronessee} & 2022  & 5-260  & Maritime   & 14K & 60K \\  
  MAVREC~\cite{MAVREC} & 2024  &  25–45  & Various   & 14K & 60K \\  
   
\bottomrule
\end{tabular}}

\end{table}

\vspace{-6px}
\subsection{Approaches for aerial human detection}
The choice of network architecture is crucial for all deep-learning-based approaches \cite{HumanDetectionUAVsurvey}. The aerial human detection community has adopted various network architectures, including two-stage network architecture \cite{DSOD,AdaptiveAnchor,FusionFactor}, one-stage network architecture \cite{DSHNet,FusionFactor,QueryDet}, anchor-free architecture \cite{PENet,RRNet}, and network ensembles \cite{VisDroneDET2020,VisDroneDET2020}.

\vspace{3px}
\noindent \textit{Two-stage networks:}
Due to its performance, two-stage detectors such as Faster R-CNN \cite{FasterRCNN} are popular in the literature \cite{DMNet,PLAOD,GLSA,FusionFactor}. In aerial object detection, a strong multi-scale representation is crucial due to the variance of object sizes and resolutions, which reflects the popularity of the FPN backbone \cite{DSOD,AdaptiveAnchor,FusionFactor}. The multi-scale challenge also makes multi-stage detectors well-suitable for aerial object detection. Many recent approaches \cite{SAMR,SyNet,GDFNet} achieved good performance with Cascade R-CNN \cite{CascadeRCNN}. In fact, many entries including the winning and runner-up entries, of the VisDrone detection challenges combined Cascade R-CNN \cite{CascadeRCNN} with other networks to achieve high detection performance \cite{VisDroneDET2020}. Similarly, two of the top three detectors in the TinyPersons challenge employed Cascade R-CNN \cite{TinyPersonsChallenge}.

\vspace{3px}
\noindent \textit{One-stage networks:}
One-stage networks are fast and less computationally intensive than their multi-stage counterparts, hence they have also been employed in many aerial object detection approaches. RetinaNet \cite{RetinaNet} is among the most popular one-stage networks for aerial object detection by \cite{DSHNet,FusionFactor,QueryDet}. EfficientDet \cite{EfficientDet} has also been employed in the aerial setting by \cite{ScaleInvarianceAerial,DomainLabels}. Real-time detectors such as YOLO and its variants are also a popular one-stage network for aerial object detection by \cite{GLSA,SelectiveTiling}. Pelee \cite{Pelee}, a real-time object detection system on mobile devices, has also been utilized in aerial object detection by \cite{TilingPower}.

\vspace{3px}
\noindent \textit{Anchor-free:}
In aerial images where object instances vary drastically in resolution and size in a single image, finding good prior anchor sizes is not feasible. LaLonde \emph{et al.} \cite{ClusterNet} also showed that it is easier to locate extremely small objects with points instead of anchors, hence many approaches have shifted to using anchor-free network designs. CenterNet \cite{CenterNet} with an Hourglass backbone \cite{Hourglass} has been a popular choice, such as \cite{AdaptiveSearching,AdaptiveFeat}. \cite{SODA} took advantages of FCOS \cite{FCOS,FCOSPAMI}. Tang \emph{et al.} \cite{PENet} employed a CornerNet design \cite{CornerNet}. Zhang \emph{et al.} \cite{GDFNet} employed a FreeAnchor design \cite{FreeAnchor}. Chen \emph{et al.} \cite{RRNet} also designed a point-based detector called RRNet and observed that point-based detectors usually outperform all anchor-based detectors in the VisDrone test set.

\vspace{3px}
\noindent \textit{Network ensemble:}
Each detector may have different strengths and weaknesses. While the multi-stage detectors tend to produce more false negatives, which means that multi-stage detectors fail to detect some objects, single-stage detectors generally propose more bounding boxes with lower quality \cite{SyNet,ObjectDetection20Years}. Hence, combining them may predict more bounding boxes than multi-stage detectors, and the quality of the single-stage detector predictions may be enhanced by the multi-stage one. Albaba \emph{et al.} \cite{SyNet} showed that combining a multi-stage detector, \emph{i.e.} Cascade R-CNN \cite{CascadeRCNN}, and a single-stage detector, \emph{i.e.} CenterNet \cite{CenterNet}, yields higher accuracy than individual detectors. The ensemble strategy has been widely employed in practice since it is effective in improving the accuracy of object detection. For example, the winner of the VisDrone object detection challenge 2020 \cite{VisDrones}, DPNetV3 \cite{VisDroneDET2020}, ensembles a few powerful backbones such as HRNet-W40 \cite{HRNet}, Res2Net \cite{Res2Net}, Balanced Feature Pyramid Network \cite{LibraRCNN} and Cascade R-CNN \cite{CascadeRCNN}. The second-ranked detector also uses different combinations of multiple models (\emph{i.e.} Cascade R-CNN \cite{CascadeRCNN}, HRNet \cite{HRNet}, and ATSS \cite{ATSS}).


\vspace{-6px}
\subsection{Techniques to solve aerial detection challenges}
The distinct aerial challenges discussed in Section~\ref{sec:AerialHumanDetChallenges} arise in a large number of domains, across which an effective aerial detection model has to stay robust. 

\vspace{-6px}
\subsubsection{Low resolution} 
The most challenging factor in aerial object detection is the low resolution or small size of objects due to the high flying altitude. A human may appear as tiny as a few pixels in an image. For example, the size of humans in the TinyPerson dataset  \cite{TinyPersons} ranges $[2,32]$, in practice $[2,20]$ is considered as tiny and $[20,32]$ is considered as small. In the VisDrone dataset, human sizes only range from 0.00014\% to 5.59\% and a mean of 0.044\% and the human objects occupy only three pixels in images within a frame resolution of $1,916\times1,078$ pixels \cite{VisDrones}.
The scale of object resolutions also varies largely from $10^1$ to $10^3$ pixels not just within the dataset, but also within a single image \cite{RRNet}. There are three key approaches to deal with small object detection in aerial data: (i) improve feature maps for small objects \cite{SmallDetSurvey,SmallDetSurvey2}, (ii) incorporate context information of small objects \cite{ContextAerial,ContextRole}, and (iii) data augmentation \cite{VisDroneDET2020}.

\vspace{-6px}
\subsubsection{Multi-scale detection} 
Multi-scale detection is commonly used in aerial object detection due to the dominance of small objects and the co-presence of object instances with a wide range of scales. Similar to generic multi-scale object detection, many aerial object detectors \cite{DSOD,PLAOD,USFA,SAMR} employ FPN \cite{FPN} and its variants. New techniques have also been proposed to improve the fusion between feature maps. Gong \emph{et al.} \cite{FusionFactor} introduced a fusion factor to weigh the aggregation between adjacent layers. Gong \emph{et al.} \cite{AdaptiveFeat} also proposed an adaptive feature selection scheme to aggregate feature maps. Wang \emph{et al.} \cite{SAMR} proposed to refine multi-scale features by increasing the receptive field size for high-level semantic features. Differently, Yang \emph{et al.} proposed a sequential approach to employ local high-resolution features for small objects in their QueryDet \cite{QueryDet}. Liu \emph{et al.} proposed to combine both feature pyramid and image pyramid to improve the aggregated features. Lingjie \emph{et al.} combined a multiscale feature extraction module with one additional small object detection head \cite{MFFSODNet}.

\vspace{-3px}
\subsubsection{Viewing angle}
Extreme views due to the high altitude of cameras, birds-eye views, and highly angled views make objects appear significantly different from popular ground-based data. Objects, \emph{i.e.} human, in aerial images may only have a top view and be arbitrarily rotated \cite{ArbitraryOrientedDetection}. Most networks can learn to cope with these large variations if the training data contains these novel views and rotations. However, this requires a large-scale dataset to cover all possible variations. Data augmentation \cite{VisDroneDET2020,TinyPersonsChallenge} and generative models such as GANs \cite{GANdetection} can be used to generate synthetic data with simulated top view and angle view to introduce these aerial viewing challenges to the training explicitly. 

\vspace{-3px}
\subsubsection{Non-uniform distribution} 
The aerial community has departed from the uniform cropping strategy \cite{TilingPower,SelectiveTiling} by learning density-specific regions to perform detection in parallel with detection on the original image.  Li \emph{et al.} \cite{DMNet} designed a multi-column CNN \cite{MCNN} to predict a density map and utilized a sliding window on the density map to generate proposal regions for cropping. Zhang \emph{et al.} \cite{FullyExploitAerial} designed a Difficult Region Estimation Network to estimate the regions that contain difficult targets and utilized a sliding window on the difficulty map to generate proposal regions for cropping. Yang \emph{et al.} \cite{ClusterDet} designed a Cluster Proposal Network (CPNet) and Tang \emph{et al.} \cite{PENet} designed a coarse network (CPEN) to predict cluster chips that have similar object scales.


\vspace{-6px}
\subsection{Techniques to improve aerial human detection}
\noindent \textit{Context modeling:} While feature pyramid networks have the capacity to aggregate multi-scale features to deal with the small and tiny object instances in aerial data, context information can provide strong cues on the presence of humans \cite{ContextRole}. DBNet \cite{VisDroneDET2020} added a global context block to improve detection. Contextual information can also be captured by dilated convolution \cite{MERI}. Similarly, Du \emph{et al.} designed a context-enhanced group normalization (CE-GN) layer by replacing the statistics based on sparsely sampled features with the global contextual ones \cite{FSAF}.

\vspace{3px}
\noindent \textit{Auxiliary meta-data:} Wu \emph{et al.} \cite{NDFT} proposed to incorporate UAV-specific nuisance annotations, which are freely available as meta-data in aerial data. For example, for UAVDT \cite{UAVDT} and VisDrones \cite{VisDrones}, three nuisance annotations (altitude, weather, and view angle) are used simultaneously with object classes to train the representation network by combining an object detection loss and a nuisance prediction loss. Kiefer \emph{et al.} \cite{DomainLabels} proposed a multi-domain strategy to leverage those nuisance annotations. 
The multi-domain strategy improves representation, which subsequently boosts detection accuracy on both datasets.

\vspace{3px}
\noindent \textit{Domain Generalization:} Domain Generalization (DG) can significantly enhance aerial human detection by addressing the challenges posed by domain shifts from ground to aerial. Kunyu \emph{et al.}  \cite{DG-UAVOD} proposed a Frequency Domain Disentanglement method, which involves separating domain-invariant and domain-specific features in the frequency domain. By training the detection model on domain-invariant features, it can generalize better to unseen environments. Using synthetic data to augment training datasets can help improve the model's robustness \cite{10423161}. Techniques like sim2real \cite{lee2024exploringimpactsyntheticdata} transformation can bridge the gap between synthetic and real-world data, enhancing the model's ability to generalize across ground and aerial domains.

\vspace{-6px}
\subsection{Aerial human detection beyond visible}
Thermal imaging plays a critical role in aerial surveillance, particularly in low-light, nighttime, and adverse weather conditions where RGB sensors often fail. Several adaptations of popular object detectors have demonstrated promising results on thermal datasets. YOLO-Thermal \cite{ALSS-YOLO}, a variant of YOLOv8 retrained on thermal imagery, has shown competitive accuracy and superior speed in detecting humans under challenging conditions such as fog and rain. Similarly, Faster R-CNN and SSD have been fine-tuned on aerial thermal datasets, with Faster R-CNN using ResNet50 achieving high detection accuracy and SSD with MobileNet-v1 offering real-time performance \cite{electronics11071151,DetectionThermalFCRN}. Schedl  \emph{et al.} further demonstrated that aerial thermal person detection under occlusion conditions can be notably improved by combining multi-perspective images before classification, achieving precision and recall rates of 96\% and 93\%, respectively, using synthetic aperture imaging techniques \cite{SearchRescureThermal}.

Beyond single-modality detection, sensor fusion strategies that combine RGB and thermal data have emerged as powerful tools to enhance robustness. These fusion methods are typically categorized into early fusion (input-level), mid fusion (feature-level), and late fusion (decision-level), depending on the stage at which modalities are integrated. Early fusion allows joint representation learning but requires careful alignment, while late fusion offers flexibility by merging outputs from independent detectors. Recent works such as RRNet \cite{RRNet} and COMET \cite{marvasti2020comet} have explored dynamic fusion architectures that adaptively weigh the contribution of each modality based on environmental context, significantly improving detection reliability in real-world aerial surveillance scenarios.


\section{Aerial Human Identification}  
\label{sec:HumanIdentification}
Person recognition is of paramount importance to human-centric aerial surveillance. To identify humans, biological biometric traits, \emph{e.g.} face, periocular, iris, fingerprint, or behavioral biometric traits, \emph{e.g.} gait, keystroke, voice, signature, cognitive, have been investigated. Aerial human identification is emerging quickly as an important area of research as evidenced by the Biometric Recognition and Identification at Altitude and Range (BRIAR) program from the Intelligence Advanced Research Projects Activity (IARPA) \cite{BRIAR}. The BRIAR program aims to identify or recognize individuals at long-range (\emph{e.g.} 300+ meters), through atmospheric turbulence, or from elevated and/or aerial sensor platforms (\emph{e.g.} $>20^o$ sensor view angle from watch towers or UAV). However, due to the unique characteristics of aerial footage, human identification from aerial footage is very challenging, even for humans \cite{PersonReIDUAVsurvey}. This section reviews aerial face, gait, and wholebody recognition.

\vspace{-6px}
\subsection{Aerial Face Recognition}
\label{sec:facerecog}
Aerial face recognition is a critical component of human-centric aerial surveillance, enabling the identification of individuals from airborne platforms. This task addresses the question: Who is the person being observed from above? Unlike ground-based face recognition, aerial scenarios introduce extreme challenges due to high altitudes, long-range imaging, and non-cooperative subjects. Faces often appear at low resolution, under varying illumination, and from unconventional angles, making reliable recognition significantly more difficult. Despite its potential utility in applications such as search and rescue, border security, and crowd monitoring, aerial face recognition remains one of the most technically demanding tasks in the aerial surveillance domain.

\vspace{-3px}
\subsubsection{Challenges for aerial face recognition}
\begin{figure}
    \centering
    \includegraphics[width=0.8\columnwidth]{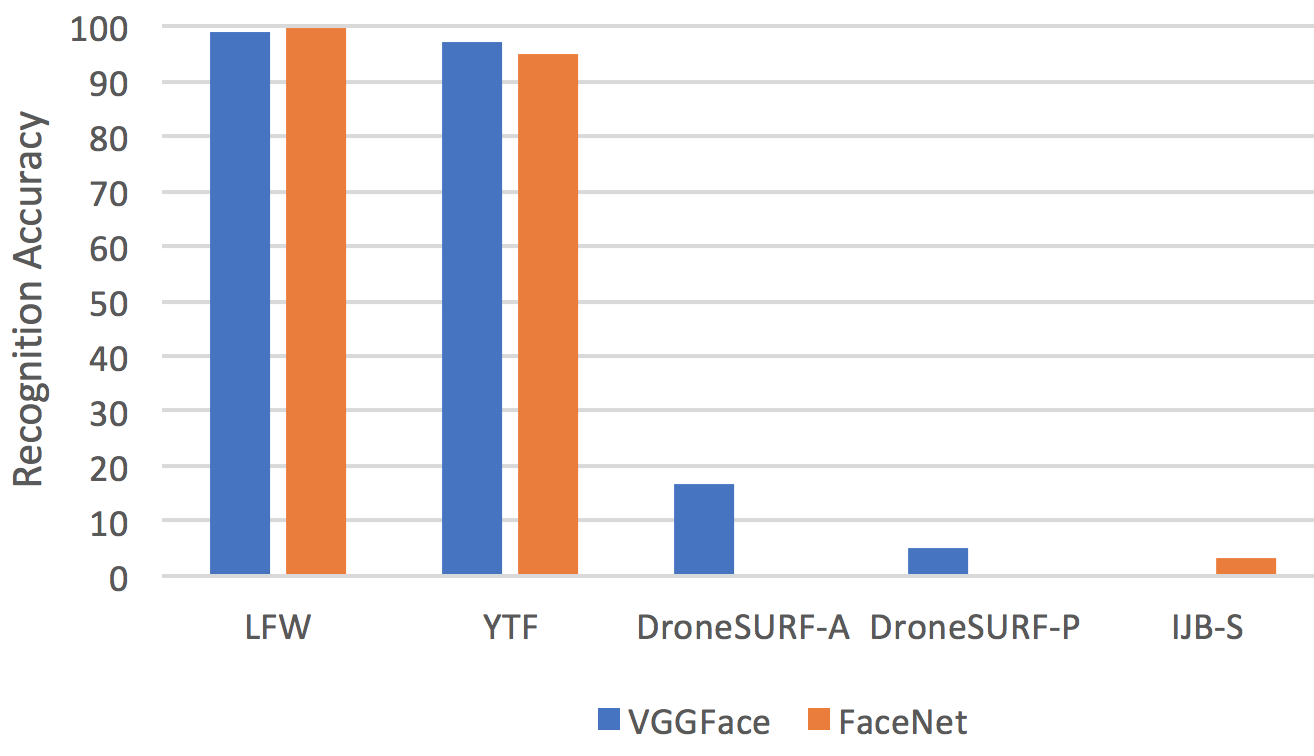}
    \caption{State-of-the-art generic face recognizers drop the performance significantly when shifting to the aerial data. The figure compares the accuracy of VGGFace \cite{VGGFace} and FaceNet \cite{FaceNet} on two ground-based (LFW and YTF) and two aerial (DroneSURF (Active and Passive) and IJB-S) datasets.}
    \label{fig:AerialFRPerformance}
\end{figure}

Despite the great success of the generic face recognition methods trained and tested on ground images, an enormous performance drop is observed when they are directly applied to images captured by UAVs \cite{DroneSURF,E2EunconstrainedFR}. Compared with other tasks, aerial face recognition may exhibit the most drop in performance when migrating from ground to aerial  data. Examples of the performance drop of state-of-the-art face recognizers are illustrated in Fig.~\ref{fig:AerialFRPerformance}. The VGGFace \cite{VGGFace} reduces the accuracy of face recognition from 99.13\% in a ground dataset (LFW) to 16.78\% in an aerial dataset (DroneSURF-Active) where subjects are cooperative and 4.95\% in a dataset (DroneSURF-Passive) where subjects are covertly surveilled \cite{DroneSURF}. The FaceNet \cite{FaceNet} drastically dropped its performance from 99.63\% in a ground dataset (LFW) to 3.33\% in an aerial dataset (IJB-S) \cite{E2EunconstrainedFR}.

The unsatisfactory performance is owing to the domain shift caused by the high flying altitude and the camera characteristics. The small size of a face itself \cite{LRface,SRface} tends to amplify all seven challenges discussed in Section~\ref{sec:AdvantagesChallenges}, making aerial face recognition extremely challenging.

\subsubsection{Datasets for aerial face recognition}
The number of public datasets and research in aerial face recognition is extremely limited, likely (at least partially) attributed to privacy concerns. Among public datasets available as illustrated in Table~\ref{tab:FRDatasets}, two noticeable ones are DroneSURF \cite{DroneSURF} and BRIAR \cite{BRIARdataset}. 

\begin{figure}
    \centering
    \includegraphics[width=\columnwidth]{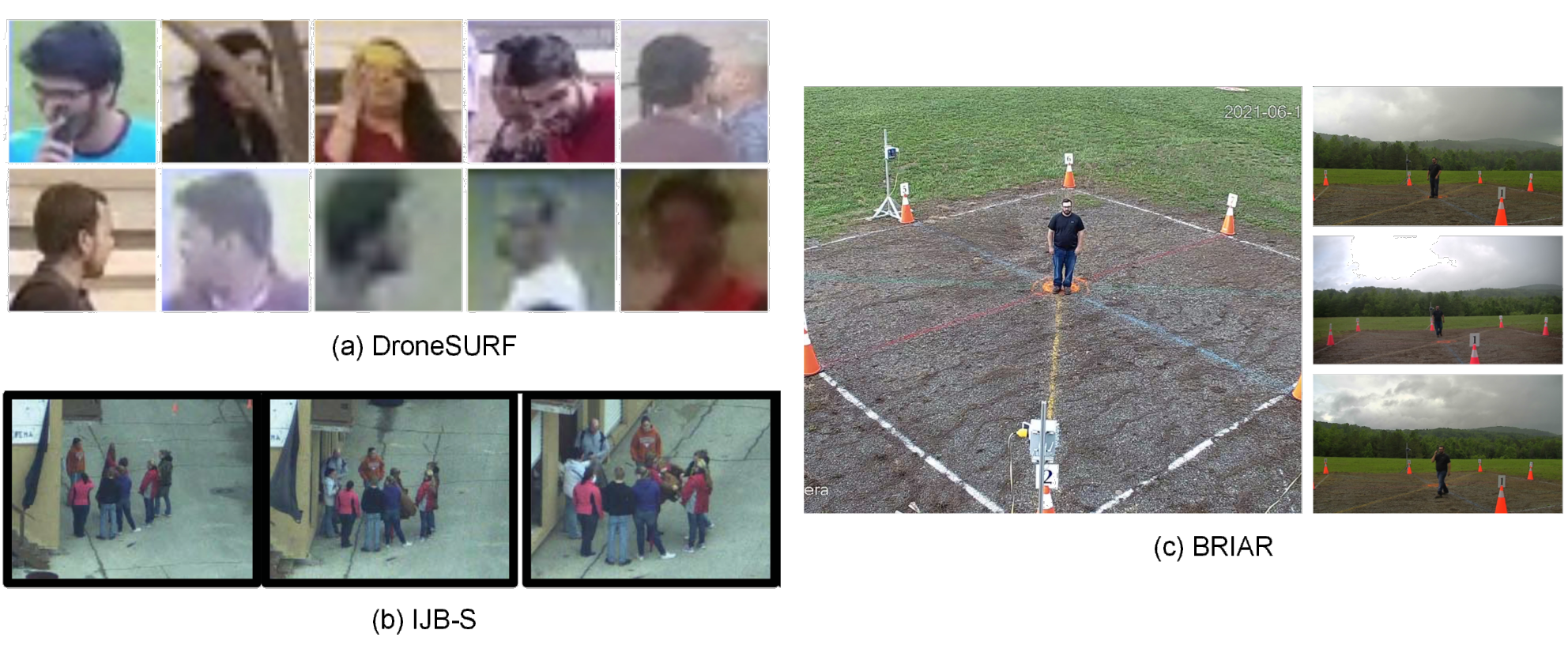}
    \caption{Examples of aerial facial images from three datasets: (a) DroneSURF \cite{DroneSURF}, (b) IJB-S \cite{IJB-S} and (c) BRIAR \cite{BRIARdataset}.}
    \label{fig:Aerial_FR_Datasets}
\end{figure}

\begin{itemize}
    \item \textit{DroneSURF 2019} \cite{DroneSURF} is the most realistic dataset for aerial face recognition. The dataset demonstrates variations across two surveillance use cases: (i) active and (ii) passive collaboration from participants. It contains 200 videos of 58 subjects, captured across 411K frames, having over 786K face annotations. DroneSURF encapsulates challenges due to the effect of motion, variations in pose, illumination, background, altitude, and resolution, especially due to the large and varying distance between the drone and the subjects. 
    \item \textit{BRIAR 2021} \cite{BRIARdataset} is a large-scale dataset for face and whole-body recognition at altitude and range, comprising over 350,000 images and 1,300 hours of video from approximately 1,000 subjects. Data were collected using UAVs (up to 400 meters) and ground setups across diverse terrains and lighting conditions, simulating real-world operational environments. Examples of images from three datasets are shown in Fig.~\ref{fig:Aerial_FR_Datasets}.
\end{itemize}

\begin{table}[t]
\small
\caption{Public datasets for aerial face recognition. `Con.', `\#ID', `\#Frm', and `\#Vid.' respectively represent the context where data was collected, the number of identities, frames, and videos.}
\label{tab:FRDatasets}
\centering
\resizebox{\columnwidth}{!}{
\begin{tabular}{l l l c c r r r r}\toprule
 {Dataset}~~ &  {Year} & {Alt.} & {Con.} & {\#ID.} & {\#Frm} & {\#Vid.}\\
\midrule
\midrule
  DroneFace~\cite{DroneFace}  & 2017   & 1-5m  & Campus   & 11 & 2,057 & -  \\
  IJB-S~\cite{IJB-S}          & 2018    & 10m  & Marketplace & 202 & 632 & 10  \\
  DroneSURF~\cite{DroneSURF}  & 2019    & -  & Various & 58 & 411K & 200\\
  BRIAR~\cite{BRIARdataset}   & 2023    & 0-400m  & Various & 1000 & 350K & - \\
\bottomrule
\end{tabular}}
\end{table}

\subsubsection{Approaches for aerial face recognition}
Surprisingly the number of published works on aerial face recognition is scarce. There is only one paper on the DroneFace \cite{DBFR}, one paper on the DroneSURF \cite{MRFR}, and four papers on the IJB-S-UAV \cite{E2EunconstrainedFR,CFAN,MARN,REAN}. While targeting aerial facial recognition, all of these had no mechanism to deal with the unique challenges of aerial footage. For example, both \cite{DBFR} and \cite{MRFR} simply employed a classification approach with a cross-entropy loss to classify a probe face into a closed gallery list. All work \cite{CFAN,MARN,REAN} on the IJB-S-UAV simply investigated different strategies to aggregate frames in a video. However, these early attempts have shown the difficulty of aerial facial recognition. For the IJB-S-UAV dataset, the best model \cite{MARN} only achieved a rank-1 accuracy of 7.63\% in a closed-set setting and 3.13\% in an open-set setting. For the DroneSURF dataset, the best model \cite{MRFR} only achieved recognition accuracy of 24.25\% on the active use case and 3.72\% on the passive use case. When probe images are carefully cropped, the accuracies can reach 60.87\% on the active use case and 45.84\% on the passive use case.
Recent advancements include the work, AerialFace~\cite{AerialFace}, which introduces a novel light-weight UAV face recognition framework. Employing the Residual SRGAN to enhance image quality and the Semantic-improved MobileFaceNet for accurate recognition amidst complex backgrounds, this framework is tailored for UAVs' limited computational capacities.



\vspace{-6px}
\subsection{Aerial Gait Recognition}
\label{sec:gaitrecog}
Aerial gait recognition focuses on identifying individuals based on their walking patterns captured from airborne platforms. This task addresses the question: How can we recognize a person from their movement in aerial footage? Unlike static biometric traits such as facial features, gait offers a dynamic and non-intrusive modality that can operate at a distance without subject cooperation. However, aerial settings introduce significant challenges including low resolution, extreme viewing angles, and frequent occlusions. These factors make gait recognition from UAVs particularly complex, yet promising for applications in long-range identification, persistent tracking, and behavioral analysis.

\subsubsection{Challenges for aerial gait recognition}
\begin{figure}
    \centering
    \includegraphics[width=1\columnwidth]{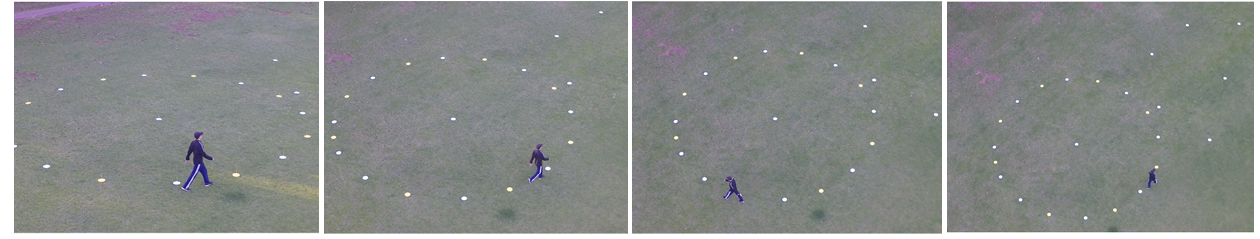}
    \caption{Challenges in aerial gait recognition: large tilt angles and low resolution.} 
    \label{fig:Aerial_Gait_Datasets}
\end{figure}

Despite the great progress of generic gait recognition methods trained on ground-to-ground images, the performance of a ground-data-trained model often degrades on aerial data due to the domain shift issue caused by the large discrepancy of pose, resolution and clothing appearance (as shown in Fig.~\ref{fig:Aerial_Gait_Datasets}).

\vspace{3px}
\noindent\textbf{Large pose discrepancy}: Most gallery gaits are captured with ground cameras while the probe gaits are from aerial views with large tilt angles, which makes gait recognition challenging.

\vspace{3px}
\noindent\textbf{Low image resolution}: Given the altitude of the aerial sensor, the probe gait videos are often captured at lower image resolutions, which diminishes the gait information. 

\vspace{3px}
\noindent\textbf{Changing clothing}: This is the classic challenge for gait recognition where the subject in the gallery and probe could wear completely different clothing, resulting in a large appearance discrepancy. 

\begin{table}[t]
\small
\caption{Public datasets for aerial gait recognition. `Alt.', `\#ID', `\#Frm.' and  `\#Cam.' respectively represent altitude, number of subjects, number of sequences, and number of cameras.}
\label{tab:aerialGaitDatasets}
\centering
\resizebox{\columnwidth}{!}{
\begin{tabular}{l l l c c r r r r}\toprule
 {Dataset}~~ &  {Year} & {Alt.} & {Context} & {\#ID.} & {\#Seq} & {\#Cam.}\\
\midrule
\midrule
  UAV-Gait~\cite{UAV-Gait}      & 2022      & 15-45  & Campus & 1,615 & 9,898  & 3\\
  DroneGait~\cite{DroneGait}    & 2023      & 20-60  & Various & 96 &  22,718  & 10\\  
  AerialGait~\cite{AerialGait}  & 2024      &  3-20  & Synthetic & 533 & 82,454  & -\\  
\bottomrule
\end{tabular}}
\end{table}

\subsubsection{Datasets for aerial gait recognition}
We summarize public datasets and their statistics in
Table ~\ref{tab:aerialGaitDatasets}.

\begin{itemize}
    \item \textit{DroneGait}~\cite{DroneGait} is a dataset designed for gait recognition from high vertical views, featuring 22,718 sequences from 96 subjects captured at vertical angles ranging from 0° to 80°. The dataset includes diverse conditions such as clothing variations, carried items, and changing lighting. It offers multiple data modalities, including silhouettes, 2D/3D poses, 3D meshes, and optical flow, making it suitable for appearance-based and skeleton-based models. 
    \item \textit{AerialGait} \cite{AerialGait} is a large-scale dataset for aerial-ground gait recognition, comprising 82,454 sequences from 533 subjects across five surveillance sites. Data were captured using DJI Mavic 3 drones at varying altitudes (3–20m), speeds (1–15m/s), and flight paths, along with ground cameras. The dataset includes annotated silhouettes, 2D/3D poses, and parsing data, and reflects challenges such as view variation, motion blur, and domain gaps between aerial and ground perspectives.
\end{itemize}

\subsubsection{Approaches for aerial gait recognition}

There are very few existing gait recognition methods specifically working in the aerial domain.
\cite{sien2019deep} proposes a two-step framework that first detects humans in a video based on a Single-Shot Multi-box Detection (SSD), and then leverages an LSTM-based recurrent processing structure to perform gait recognition.
\cite{perera2018human} estimates the gait sequence and movement trajectory of a person from an aerial video. 
The proposed solution consists of a perspective correction module, feature extraction module, pose estimation module, and trajectory estimation module. 
Ding \emph{et al.}~\cite{UAV-Gait} enhances the GaitSet framework by introducing a graph convolution-based part feature pooling module to address feature misalignment caused by pitch rotations in UAV-based gait recognition. By aggregating spatially and temporally adjacent part features using graph convolution, the method effectively captures local body shape details, improving robustness to large view variations. 
V-Distill~\cite{DroneGait} introduces a dual-branch feature distillation framework, aiming to align features between moderate (low to middle) and high vertical views using a pre-trained branch for moderate views and a learnable branch for high views. By projecting high-angle features into a discriminative space, V-Distill addresses challenges like occlusion and reduced resolution.
AGG-Net~\cite{AerialGait} addresses several challenges unique to aerial gait recognition, such as significant view variations, motion blur, and domain gaps between aerial and ground views. It comprises two key components: a gait-oriented Uncertainty Learning module, which introduces uncertainty at the silhouette and feature levels to enhance robustness against noise and variations in input data, and an aerial-ground prototype learning module, which aligns feature distributions between aerial and ground views through prototype clustering.

\vspace{-6px}
\subsection{Aerial Whole-Body Recognition}
\label{sec:wholebodyrecog}
Whole-body recognition, which integrates multiple biometric modalities such as face, body shape, and gait, offers a comprehensive approach for identifying individuals from a distance. This multimodal biometric system is particularly advantageous in aerial scenarios, where different characteristics may be visible depending on the angle, distance, and quality of the capture. The dynamic information gathered from aerial perspectives presents unique challenges but also significant opportunities for enhancing surveillance and monitoring capabilities.

\subsubsection{Challenges for aerial whole-body recognition}
The extension of whole-body recognition to aerial platforms introduces a unique set of challenges that complicate the identification process:

\vspace{3px}
\noindent\textbf{Varying angles and distances}: Aerial platforms often capture images and videos from high altitudes and varied angles, introducing significant variations in body orientation and scale that can obscure key biometric features.

\vspace{3px}
\noindent\textbf{Environmental factors}: Factors such as lighting, weather, and atmospheric conditions can severely affect the visibility and quality of captured biometric data, influencing the performance of recognition systems. 

\vspace{3px}
\noindent\textbf{Multimodal fusion challenges}: Integrating features from face, body, and gait requires robust algorithms capable of handling discrepancies in data quality and resolution, which are common in aerial captures. The fusion must be adaptive to account for the varying reliability of different biometric modalities at different distances and viewing angles.

\subsubsection{Datasets for aerial whole-body recognition}

The BRIAR dataset~\cite{BRIAR} encompasses a wide range of environmental, operational, and subject-based variabilities. Specifically designed to address the unique challenges of aerial and long-range biometric recognition, BRIAR includes diverse scenarios featuring individuals at various distances and altitudes. This dataset is rich with imagery captured from UAVs, providing data that include different angles, poses, and motion states of subjects. It offers a realistic set of conditions that algorithms must overcome, such as variable lighting, partial occlusions, and different weather conditions, making it an invaluable resource for developing robust aerial whole-body recognition systems.

\subsubsection{Approaches for aerial whole-body recognition}

Increasingly, researchers are recognizing the importance of innovative methodologies that are shaping the future of aerial whole-body recognition, enhancing system capabilities to accurately identify individuals from challenging aerial perspectives. 
FarSight~\cite{FarSight}
Introduces a system that effectively combines physics-based modeling with deep learning to enhance the resolution and biometric feature extraction of images captured at long distances and high altitudes. This system focuses on overcoming the challenges of low-resolution and large domain gaps typical in aerial images.
CoNAN~\cite{CoNAN} describes a network that uses a novel feature aggregation technique to handle biometric data captured under highly unconstrained settings. It significantly improves the fusion of biometric features such as face and body shapes, enhancing recognition performance in aerial scenarios.
Complementarily, an in-depth analysis~\cite{bolme2024data} of the BRIAR dataset identifies key environmental and operational factors affecting recognition accuracy, guiding further refinement of recognition algorithms. 
%
Additionally, BRIARNet~\cite{BRIARNet} offers a methodology for robust detection and identification of individuals using whole-body biometrics from high altitudes. The paper outlines an end-to-end solution for integrating and optimizing detection and recognition processes specific to challenging aerial data.
Looking ahead, the field of aerial whole-body recognition is poised for transformative advancements as the integration of AI and drone technologies continues to evolve. Future research will likely explore more sophisticated algorithms that can dynamically adapt to rapid changes in environmental conditions and movement dynamics of subjects. Another promising area is the development of more advanced sensor technologies that can capture higher resolution data from greater distances, further improving the reliability and accuracy of recognition systems. Additionally, as privacy concerns grow, new methods that can secure biometric data while maintaining high accuracy in identification will become crucial. The ultimate goal will be to create fully autonomous systems that can operate in a variety of complex environments, providing reliable surveillance and identification capabilities that are crucial for both security and humanitarian applications.

\section{Aerial Person Re-Identification}
\label{sec:HumanReID}
Aerial person re-identification (Re-ID) is a critical task in human-centric aerial surveillance, focusing on the retrieval of individuals across multiple non-overlapping cameras. This task addresses the fundamental questions: Where and when has this person been seen in the surveillance network? The query individual can be represented through visual cues such as images or videos, or textual descriptions \cite{ReIDsurvey}. Aerial person Re-ID is particularly practical in unconstrained surveillance environments, where high-flying cameras and uncooperative subjects complicate the precise measurement of biometric traits like the face. Despite its practicality, aerial person Re-ID faces unique challenges due to the distinct characteristics of aerial footage, including varying altitudes, extreme viewing angles, and low-resolution imagery. Addressing these challenges is essential for enhancing the accuracy and reliability of aerial person Re-ID systems.

\subsection{Challenges for aerial person re-ID}
\begin{figure}
    \centering
    \includegraphics[width=0.8\columnwidth]{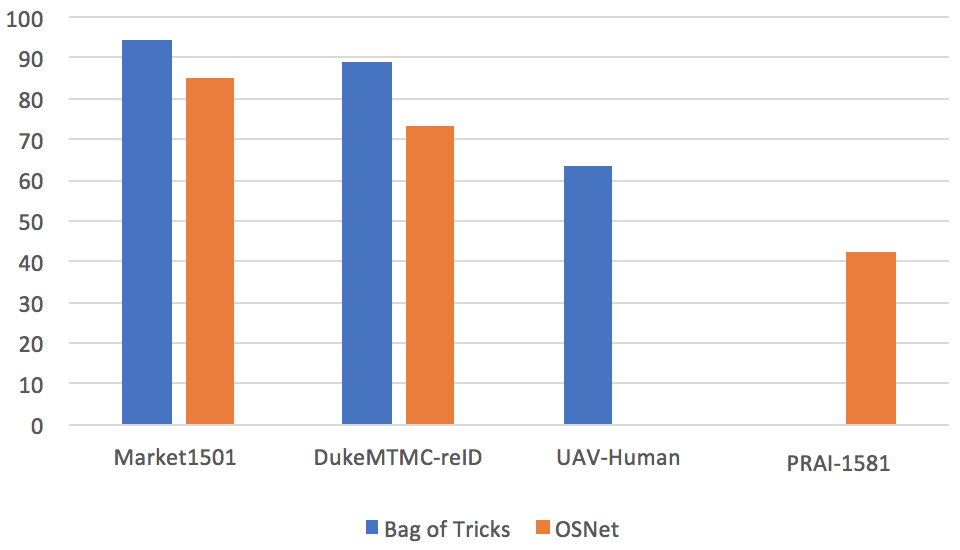}
    \caption{State-of-the-art generic person re-ID algorithms drop the performance when shifting to the aerial data. The figure compares the re-identification accuracy of Bag of Tricks \cite{BoT} and OSNet \cite{OSNet} on two ground-based (Market1501 and DukeMTMC) and two aerial (UAV-Human and PRAI-1581) datasets.}
    \label{fig:AerialReIDPerformanceDrop}
\end{figure}

Compared with other tasks, person re-id may exhibit the least drop in performance when migrating from ground to aerial data. Examples of the performance drop of state-of-the-art person re-id algorithms are illustrated in Fig.~\ref{fig:AerialReIDPerformanceDrop}. The state-of-the-art Bag of Tricks \cite{BoT} reduces the accuracy from 94.2\% and 89.1\% in two ground datasets, Market1501 and DukeMTMC, to 63.4\% in the aerial UAV-Human dataset \cite{UAV-Human}. Similarly, the OSNet \cite{OSNet} reduces the accuracy from 84.9\% and 73.5\% in two ground datasets, Market1501 and DukeMTMC, to 42.1\% in the aerial PRAI-1581 dataset \cite{PRAI1581}.

\begin{figure}
    \centering
    \includegraphics[width=\columnwidth]{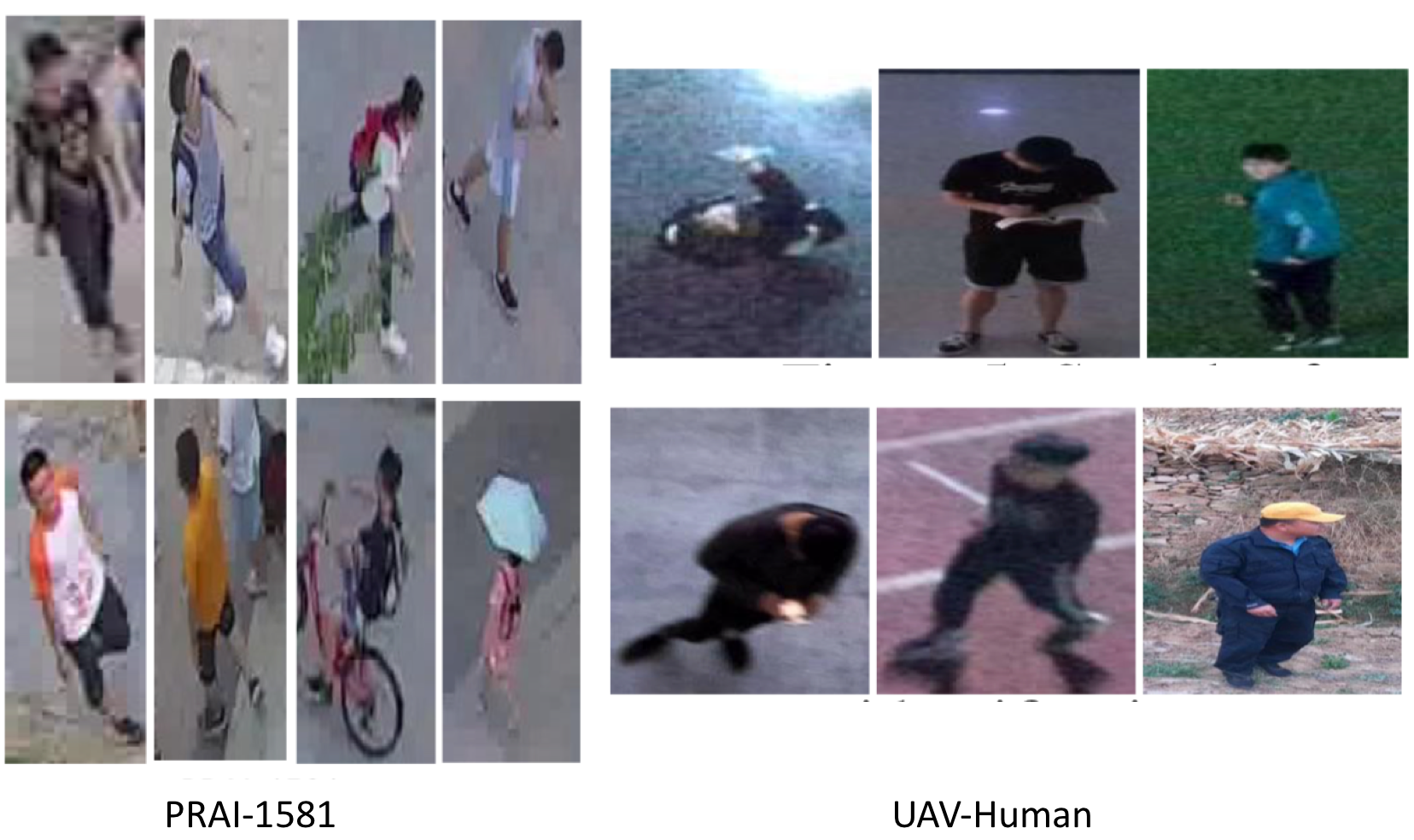}
    \caption{Challenges for aerial person re-ID due to elevation angle, blur, lighting/illumination, and occlusion. Images from two aerial person re-ID datasets: PRAI-1581 \cite{PRAI1581} and UAV-Human \cite{UAV-Human}.}
    \label{fig:AerialPersonReID_Datasets}
\end{figure}

The performance gap is owing to the domain shift caused by high-flying altitudes and camera characteristics. Aerial Person Re-ID exhibits all seven challenges discussed in Section~\ref{sec:AdvantagesChallenges}. Exemplar challenges are illustrated in Fig.~\ref{fig:AerialPersonReID_Datasets}.

\begin{table}[t]
\small
\caption{Public datasets for aerial person re-ID. `Alt.', `\#ID', `\#Frm.' and  `\#Cam.' respectively represent altitude, number of subjects, number of frames, and number of cameras.}
\label{tab:aerialReIDDatasets}
\centering
\resizebox{\columnwidth}{!}{
\begin{tabular}{l l l c c r r r r}\toprule
 {Dataset}~~ &  {Year} & {Alt.} & {Context} & {\#ID.} & {\#Frm} & {\#Cam.}\\
\midrule
\midrule
  MRP~\cite{MRP}       & 2015      & 10  & Campus   & 28 & 4,096 & -  \\
  p-DESTRE~\cite{PDESTRE}   & 2021      & 5-6   & Campus & 269 & 14.8M & 1\\
  PRAI-1581~\cite{PRAI1581} & 2021      & 20-60  & Various & 1,581 & 39K & 2\\
  UAV-Human~\cite{UAV-Human}& 2021      & 2-8    & Various & 1,144 & 41K & 1\\
  AG-ReID~\cite{AGReID2023} & 2023      & 15-45  & Campus & 388 & 21K  & 2\\
  AG-ReIDv2~\cite{AGReIDv2} & 2024      & 15-45  & Campus & 1,615 & 100K  & 3\\
  G2AVReID~\cite{G2AVReID}  & 2024      & 20-60  & Various & 2,788 & 180K  & 2\\  
  CARGO~\cite{CARGO}        & 2024      &  5-75  & Synthetic & 5,000 & 108K  & 13\\  
  LAGPeR~\cite{SeCap}        & 2025      &  20-60  & Campus &  4,231 &  64K  & 21\\  
  AG-VPReID~\cite{AGVPReID2025}        & 2025      &  15-120  & Campus & 6,632 & 9.6M  & 6\\  
  
\bottomrule
\end{tabular}}
\end{table}
\subsection{Datasets for aerial person re-ID}
One early dataset for aerial person re-ID is MRP collected by Layne \emph{et al.} \cite{MRP} in 2014 but their dataset has never been made public. Since then, due to the affordable availability of off-the-shelf drones with higher resolution cameras, the number of aerial datasets has quickly emerged in the last three years. We summarize the public datasets and their statistics in Table~\ref{tab:aerialReIDDatasets}. Notable large-scale datasets are PRAI-1581 \cite{PRAI1581}, UAV-Human \cite{UAV-Human}, and AG-ReID.v2 \cite{AGReIDv2}.
\begin{itemize}
    \item \textit{PRAI-1581 2021} \cite{PRAI1581}: Zhang \emph{et al.} recently collected a large dataset for aerial person re-ID. The images were shot by two DJI drones at an altitude of 20 to 60 meters. The dataset consists of 39k images of 1581 unique subjects. The resolution of persons is low, ranging from 30 to 150 pixels. The high flying altitude makes the diversity of views and poses more extreme. 
    \item \textit{UAV-Human 2021} \cite{UAV-Human}: Li \emph{et al.} flew one UAV in multiple urban and rural districts in both daytime and nighttime for three months, thus covering extensive diversities w.r.t. subjects, backgrounds, illuminations, weathers, occlusions, camera motions, and flying altitudes. It contains videos and annotations for multiple tasks, including person re-id and attribute recognition. There are 41,290 frames and 1,144 identities for person re-id and 22,263 frames for attribute recognition. The unique characteristic of the dataset is the multimodal aerial data captured using a depth sensor (Azure DK), a fisheye camera, and a night-vision camera.
    \item \textit{AG-ReID 2023-2024} \cite{AGReID2023}: Nguyen \emph{et al.} published a new setting of aerial-ground person ReID, which aims to investigate how to match persons cross-platform. The main challenges are the differences in viewing angles between aerial cameras and ground-based cameras. Starting with 388 identities with two cameras ( aerial camera \& CCTV camera) \cite{AGReID_Challenge}, they later expanded to 1,615 identities with three cameras (aerial camera \& CCTV camera \& wearable camera) \cite{AGReIDv2}. 
\end{itemize}

\vspace{3px}
\noindent\textbf{Long-term Aerial Re-ID:} p-DESTRE \cite{PDESTRE} provides footage captured over multiple days where the actors may change clothing and accessories, which can facilitate the long-term aerial re-ID task.

\vspace{3px}
\noindent\textbf{Hetero Aerial Re-ID IR:} UAV-Human \cite{UAV-Human} provides both RGB and thermal IR footage that can be used for multi-modal aerial person re-ID across RGB and IR spectrum.

\vspace{3px}
\noindent\textbf{Aerial Human Attribute Recognition:} UAV-Human \cite{UAV-Human} provides 7 human attributes (gender, hat, backpack, upper clothing color, and style, as well as lower clothing color and style). All three datasets AG-ReID, AG-ReID.v2, and p-DESTRE \cite{PDESTRE} provide 16 human attributes (gender, age, height, body volume, ethnicity, hair color, hairstyle, beard, mustache, glasses, head accessories’, ‘body accessories’, ‘action’ and ‘clothing information’ (x3)). These attributes can be used for aerial attribute recognition, person search, or to support the aerial re-ID task.

\subsection{Approaches for aerial person re-ID}

\vspace{3px}
\noindent\textbf{Closed-world Aerial Person Re-ID} \\
Closed world person re-ID, refers to the search where the person in the probe image is definitely present as one of the candidates in the gallery images. The majority of prevailing approaches for closed world person re-ID focus on domain invariant settings, \emph{i.e.} ground-ground and aerial-aerial person re-ID. Classical handcrafted features and classical distance metrics have been prevalent in the early stages of the development of closed-world person re-ID techniques.

\vspace{3px}
\noindent\textit{Handcrafted feature representation:} Before the explosion of deep learning in 2012, handcrafted features were used for aerial person re-ID. Oreifej \emph{et al.} \cite{FirstAerialPersonReID} directly estimated the similarity between two human blobs by the Earth Mover Distance (EMD). Layne \emph{et al.} \cite{MRP} classified human detections into a predefined ID list using a number of SVM variants. Schumann \emph{et al.} \cite{AerialPersonReID_CovFu} used a covariance descriptor and a geodesic distance between two covariance descriptors to measure their similarity. 

\vspace{3px}
\noindent\textit{Learnable feature representation:} All modern approaches have used CNNs and Transformers to learn representation directly from data. There are two categories of applying CNNs in person re-ID: discriminative and pairwise. Discriminative approaches classify each human detection into a pre-defined ID list. Schumann \emph{et al.} \cite{AerialPersonReID_Attri} designed their own CNN architecture with Inception \cite{InceptionNet} and Residual layers \cite{ResNet} to learn representation. One of the branches in the Grigorev \emph{et al.}'s framework \cite{DroneHIT} employed Resnet-50 \cite{ResNet} as a base network. They trained the base network using a large margin Gaussian mixture (L-GM) loss \cite{LGMloss}. Pairwise approaches seek to directly calculate the similarity between two human detections, eliminating the need for a pre-collected gallery list. Most approaches in the literature combine both discriminative losses and pairwise losses to train the backbone network. Grigorev \emph{et al.} \cite{DroneHIT} trained a backbone ResNet by both a triplet loss and an L-GM loss \cite{LGMloss}. Zhang \emph{et al.} also combines a triplet loss with an identification loss to train the deep network. They experimented with multiple backbone networks coupled with a subspace pooling layer to learn a compact representation \cite{PRAI1581}. Nguyen \emph{et al.} employed a Transformer architecture with a ViT backbone as the mainstream of their framework \cite{AGReID2023}. Qiu extended the self-attention mechanism of ViT by proposing a multi-branch module called Salient Part-Aware Cross-Attention (SPACA) to guide the attention to salient part features of the human body in aerial images \emph{et al.} \cite{Qiu2024}. Huang \emph{et al.} designed a multi-resolution feature perception network based on transformer architecture to handle the challenges in resolution variations in UAV scenarios, where both query and gallery images have significant resolution variations \cite{MRReID}.


\vspace{3px}
\noindent\textit{Data augmentation:} Data augmentation during training can be applied to increase the robustness against such error sources as viewpoint variations and occlusions. Moritz \emph{et al.} proposed two augmentation approaches, random rotation (RR) and random cropped rotation (RCR), to specifically improve the robustness against diverse perspectives in UAV-based person re-id \cite{ReIDdesign}. Testing with the OSNet baseline \cite{OSNet}, these augmentation techniques can boost the accuracy from 50.0\% to 53.0\% in the PRAI-1581 \cite{PRAI1581} dataset and
from 80.1\% to 83.7\% in the p-DESTRE \cite{PDESTRE} dataset.

\vspace{3px}
\noindent\textit{Transfer learning:} Many approaches \cite{AerialPersonReID_CovFu,AerialPersonReID_Attri,DroneHIT} take advantage of a large number of existing datasets in a classic ground-based setting, such as for transfer learning. A person re-ID model is first trained with these ground-based datasets, then either applied directly \cite{AerialPersonReID_CovFu} or fine-tuned \cite{AerialPersonReID_Attri,DroneHIT} on the aerial dataset. Xu \emph{et al.} proposed a meta-transfer learning strategy to enhance the feature extraction of aerial images \cite{MetaTransfer}. While this strategy has shown to be working, it is still an open question to what extent the pre-training can help, since the characteristics of aerial images and ground-based images differ. 


\vspace{3px}
\noindent\textbf{Open-world Aerial Person Re-ID} \\
Open-world person re-ID is a more difficult problem of person re-ID where the person in the probe may or may not be in the gallery. Many approaches for open-world person re-ID depart from the closed-world setting and explore aerial person re-ID across multiple modalities. This includes across domains (aerial-ground \cite{AerialPersonReID_CovFu}), across the spectrum (visible-infrared \cite{UAV-Human}), and across modalities (aerial-attribute \cite{PDESTRE}). 

\vspace{3px}
\noindent\textit{Long-term aerial person re-ID:} Long-term aerial person re-ID relaxes the time constraints between the probe and gallery imaging moment. Existing re-ID methods rely on appearance features like clothes, shoes, hair, \emph{etc.} Such features, however, can change drastically from one day to the next, leading to the inability to identify people over extended periods. The p-DESTRE dataset \cite{PDESTRE} also provides a baseline for long-term aerial person re-id based on facial and body features. The facial feature representation was obtained using the ArcFace model \cite{ArcFace} and the body feature representation was obtained using the COSAM \cite{COSAM} model. The Euclidean norm was used as a distance function between the concatenated representations. This achieved an mAP of 34.9\%, but it is noticeable that the p-DESTRE footage was captured with very low flying altitudes, \emph{i.e.} 5-6m.

\vspace{3px}
\noindent\textit{Aerial-Ground person re-ID:} 
Person re-ID between aerial images and ground-based images is of significant interest to surveillance in such tasks as large-scale search. However, the differences in human appearances, such as views, poses, and resolutions, make it very challenging. The representation has to be robust to these variations. There exists one work in aerial-ground person re-id by Schumann \emph{et al.} \cite{AerialPersonReID_CovFu} where the authors attempted to encode a robust representation via covariance descriptors. However, the views of their aerial footage and the ground-based footage are similar. In addition, the dataset is very small with only 1217 probe images and 4244 gallery images \cite{AerialPersonReID_CovFu}. Observing the attributes will be less affected by the views, Nguyen \emph{et al.} proposed an Explainable Stream to learn 15 soft attributes and leverage the heatmaps of these attributes to improve a conventional Transformer-based Re-ID stream \cite{AGReID2023}. Later, Nguyen \emph{et al.} proposed an Elevated-View Attention Stream to pay more focus on the head and shoulder regions to deal with the extreme top view of pedestrians, where only the head and shoulder may be visible \cite{AGReIDv2}. Zhang  \emph{et al.} proposed to alleviate the disruption of discriminative identity representation
by dramatic view discrepancy as the most significant challenge in Aerial-Ground PReID with a view-decoupled transformer (VDT). They designed two major components in VDT to decouple view-related and view-unrelated features, namely hierarchical subtractive separation and orthogonal loss, where the former separates these two features inside the VDT, and the latter constrains these two to be independent.

\subsection{Aerial Person Re-ID beyond Visible}

\vspace{3px}
\noindent\textit{Visible-Infrared Aerial Person Re-ID}
Surveillance is usually required to operate 24/7, where low-light or night vision, such as infrared and thermal imaging, is essential. Currently, there does not exist any work in aerial cross-spectrum person re-ID; however, it is expected to change soon with such attempts as UAV-Human \cite{UAV-Human}. Existing visible-infrared Person Re-ID in the ground-based data, \emph{e.g.} \cite{VisIRreID,VisIRreID2,VisIRreID3}, and LiDAR-based Person Re-ID \cite{LidarReID} can be experimented with to see how they perform with aerial data. New innovation is needed to solve the unique challenges of the aerial setting.

\vspace{3px}
\noindent\textit{Aerial-Attribute human recognition:}
Searching for persons in aerial data based on attribute descriptions such as gender, height, clothing, \emph{etc.} is also of practical significance in surveillance, which is imperative when the visual image of a query person cannot be obtained. The re-id task can now be compiled as a human attribute recognition problem \cite{HumanAttriSurvey}. Kumar \emph{et al.} \cite{PDESTRE} has addressed this issue by collecting the aerial dataset, p-DESTRE, and annotating persons with 16 attributes. They employed the COSAM algorithm \cite{COSAM} to search for humans with facial and body attributes, which can deal with the long-term re-id task. Nguyen \emph{et al.} also annotated 16 attributes in their datasets \cite{AGReID2023,AGReIDv2} and leveraged these attributes to improve the person re-ID results.

\begin{table*}
\caption{Performance comparison of state-of-the-art approaches in aerial face recognition, gait recognition, whole-body recognition, and person re-identification. Unless otherwise specified, the reported metric is Rank-1 accuracy. BRIAR FR denotes BRIAR FaceRestricted, BRIAR FI denotes BRIAR FaceIncluded. AG/GA denotes Aerial->Ground/Ground->Aerial setting in ReID, AC/CA denotes Aerial->CCTV/CCTV->Aerial setting in ReID. NM/BG/CL denotes Normal/Bag/Cloth settings in gait recognition.}
\label{tab:Summary}
\scriptsize
\begin{tabular}{|>{\raggedright\arraybackslash}p{0.05\linewidth}|l|l|>{\raggedright\arraybackslash}p{0.3\linewidth}|>{\raggedright\arraybackslash}p{0.08\linewidth}|>{\raggedright\arraybackslash}p{0.07\linewidth}|>{\raggedright\arraybackslash}p{0.07\linewidth}|>{\raggedright\arraybackslash}p{0.07\linewidth}|>{\raggedright\arraybackslash}p{0.07\linewidth}|}
\hline
\textbf{Recog. Task} & \textbf{Method} & \textbf{Pub.} & \textbf{Key contributions} & 
\multicolumn{4}{|c|}{\textbf{Aerial Person Recognition Datasets}}\\
\hline
& &  &  & \multicolumn{4}{|c|}{Aerial Face Recognition Datasets}\\
\cline{5-8}
\multirow{11}{*}{\parbox{0.05\linewidth}{\raggedright Aerial Face }}  &  &  &  & DroneSURF Active  & DroneSURF Passive & BRIAR FR Rank-20 & BRIAR FI Rank-20 \\
\cline{2-8}
& DroneSURF & FG'19 & Commercial-Off-The-Shelf system (COTS) & 21.88 & 8.33 &   &   \\
\cline{2-8}
& MRFR \cite{MRFR} & SSIP'20 & Fine-tuned Cross-Resolution Network   & 39.16 & 13.04 &  &  \\
\cline{2-8}
& DeriveNet \cite{DeriveNet} & T’PAMI21 & Class-specific domain knowledge losses   & 36.33 & 27.81 &  &  \\
\cline{2-8}
& SR+FR \cite{SRFR} & IJCB’24 & Super-resolution   & 51.17 &  &  &  \\
\cline{2-8}
& CoNAN \cite{CoNAN} & IJCB’23 & Conditional Neural Aggregation   & 80.21 & 13.54 &  &  \\
\cline{2-8}
& ProxyFusion \cite{ProxyFusion} & NIPS’24 & Expert Network selection   & 83.33 & 13.54 &  &  \\
\cline{2-8}
& EfficientFaceV2S \cite{EfficientFaceV2S} & ESWA’25 & EfficientNet backbones   & 61.345 & 43.02 &  &  \\
\cline{2-8}
& CR+FR \cite{CRFR} & Access’25 & Cross-resolution  & 51.55 & 26.84 &  &  \\
\cline{2-8}
& FarSight 1.0 \cite{FarSight} & WACV’24 & Physics-driven  &  &  & 26.6 & 63.6 \\
\cline{2-8}
& FarSight 2.0 \cite{FarSight2} & Arxiv’25 & Modality-specific encode, quality-guided fusion &  &  & 42.9 & 80.0 \\
\hline
&  &  & & \multicolumn{4}{|c|}{Aerial Gait Recognition Datasets}\\
\cline{5-8}
\multirow{11}{*}{\parbox{0.05\linewidth}{\raggedright Aerial Gait }} &  &  &  & DroneGait (NM,BG,CL)& AerialGait &  BRIAR FR Rank-20 & BRIAR FI Rank-20 \\
\cline{2-8}
& V-Distill \cite{DroneGait} & TMM'24 & Feature Distillation across different vertical views  & High60-80o  81.0/63.7/48.1 &  &  &  \\
\cline{2-8}
& AGG-Net \cite{AerialGait} & MM’24 & Feature clustering via Prototype Learning   & AA/GG/ AG/GA 69.6/92.3/ 54.4/48.9 & AA/GG/ AG/GA 84.9/97.3/ 76.0/80.3 &  &  \\
\cline{2-8}
& CVVNet \cite{CVVNet} & Arxiv’25 & Multi-scale Attention, Cross-Vertical-View Network   & High60-80o 34.3/36.0/36.7 Mid30-60o 88.9/91.2/85.5 &  &  &  \\
\cline{2-8}
& FarSight 1.0 \cite{FarSight} & WACV’24 & Physics-driven   &  &  & 48.6 & 49.5 \\
\cline{2-8}
& FarSight 2.0 \cite{FarSight2} & Arxiv’25 & Modality-specific encode, quality-guided fusion  &  &  & 90.6 & 93.2 \\
\hline
&  &  &  & \multicolumn{4}{|c|}{Aerial Wholebody Recognition Datasets}\\
\cline{5-8}
\multirow{6}{*}{\parbox{0.05\linewidth}{\raggedright Aerial Wholebody }} &  &   &  &   &  &  BRIAR FR Rank-20 & BRIAR FI Rank-20 \\
\cline{2-8}
& FarSight 1.0 \cite{FarSight} & WACV’24 & Physics-driven   &  &  & 62.0 & 77.7 \\
\cline{2-8}
& BRIARNet \cite{BRIARNet} & T-BIOM'24 & Pre-training and  Fine-tuning on BRIAR’s data  &   &  & 75.13 &  \\
\cline{2-8}
& FarSight 2.0 \cite{FarSight2} & Arxiv’25 & Modality-specific encode, quality-guided fusion  &  &  & 91.0 & 95.5 \\
\hline
&  &  &  & \multicolumn{4}{|c|}{Aerial Re-identification Datasets}\\
\cline{5-8}
\multirow{11}{*}{\parbox{0.05\linewidth}{\raggedright Aerial ReID}} &    &  &  & PRAI-1581 & UAV-Human &  AG-ReID (AG/GA) & AG-VPReID.v2 (AC/CA)\\
\cline{2-8}
&  RotTrans \cite{RotTrans}  & MM’22 & Rotation Invariant Transformer  & 70.8   & 75.6  &  &  \\
\cline{2-8}
& SKAKD \cite{Qiu2024}  & ICME'24  & Salient Part-Aligned and Keypoint Disentangling & 71.9 & 75.8 &  &  \\
\cline{2-8}
& SPSNet \cite{SPSNet}  & TVC’25  & Semantic-guided perspective shift   & 74.3   & 78.6  &  & 75.5  \\
\cline{2-8}
& Explainable \cite{AGReID2023} & ICME’23 & Explainable Attributes  &  &  & 81.47/82.85 &  \\
\cline{2-8}
& VDT \cite{CARGO} & CVPR'24 & View Decoupled  &   &  &  82.91/86.59 &  \\
\cline{2-8}
& V2E \cite{AGReIDv2} & TIFS'24 & Explainable Attention Network &   &  &  & 88.77/87.86 \\
\cline{2-8}
& DTST \cite{DTST} & ICME'25 & Dynamic Token Selection  &   &  & 83.48/84.72  &  \\
\cline{2-8}
& SeCap \cite{SeCap} & CVPR’25 & Self-Calibrating and Adaptive Prompts  &  &  & 84.03/76.16 & 88.12/88.24 \\
\cline{2-8}
& SD-ReID \cite{SDReID} & Arxiv’25 & View-aware Stable Diffusion  &  &  & 85.16/85.97  &  \\
\cline{2-8}
& LATex \cite{LATex} & Arxiv’25 & Attribute-based Text Knowledge  &  &  & 85.26/89.40  & 89.13/89.01 \\
\hline
\end{tabular}
\end{table*}

\section{DISCUSSION AND OPEN RESEARCH}
\label{sec:GapsOutlook}

\noindent\textbf{META ANALYSIS:} In our detailed analysis of three core surveillance tasks, we demonstrate that generic models experience a significant drop in performance when applied to aerial data. While early efforts have sought to address these challenges, results remain inconsistent across different biometric modalities. To better illustrate the current landscape, we present Table~\ref{tab:Summary}, 
which provides a comprehensive comparison of state-of-the-art methods across four key aerial person recognition tasks—face recognition, gait recognition, whole-body recognition, and person re-identification—highlighting performance gaps, dataset diversity, and methodological innovations. 

\noindent \textit{(1) Face Recognition:} Aerial face recognition remains the most challenging task, with performance dropping drastically in passive surveillance settings. Even advanced models like ProxyFusion and FarSight 2.0 show limited accuracy under extreme conditions, underscoring the need for modality-specific encoding and quality-guided fusion.

\noindent \textit{(2) Gait Recognition:}
Gait recognition shows promise, especially with models like AGG-Net and FarSight 2.0 achieving high accuracy across vertical views. However, performance varies significantly depending on view angles and dataset characteristics, indicating that view adaptation and prototype learning are critical.

\noindent \textit{(3) Whole-body Recognition:}
Whole-body recognition benefits from multimodal fusion, with FarSight 2.0 and BRIARNet demonstrating strong performance. This suggests that integrating face, body shape, and gait features can compensate for individual modality weaknesses.

\noindent \textit{(4) Person Re-identification:}
Person Re-ID exhibits the least performance drop among tasks. Transformer-based models like VDT and LATex achieve high accuracy across aerial-ground settings, showing that view decoupling and attribute-based learning are effective strategies.

Across all tasks, models that incorporate domain-specific adaptations—such as physics-informed fusion, view-aware attention, and attribute-based reasoning—consistently outperform generic baselines. This reinforces the importance of designing aerial-aware architectures rather than relying on ground-trained models. These insights from Table 5 not only benchmark current capabilities but also illuminate critical gaps and opportunities for future research, particularly in developing robust, explainable, and multimodal aerial biometric systems.

\vspace{6px}
\noindent\textbf{OPEN RESEARCH:} To advance the field, we identify four critical research directions:

\vspace{3px}
\noindent\textbf{Data:} Advancing aerial surveillance requires addressing three major data challenges: scale, diversity, and annotation.
(1) Large-scale Data: Inspired by ImageNet’s impact on vision tasks, aerial surveillance needs expansive, well-annotated datasets (e.g., VisDrones, BRIAR, UAV-Human) that reflect diverse environments and conditions.
(2) Heterogeneous Modalities: Most current datasets rely on monocular RGB imagery. Incorporating radar, LiDAR, and hyperspectral sensors can significantly improve robustness under adverse conditions like poor lighting, motion blur, and occlusion.
(3) Annotation Efficiency: Manual labeling is costly and time-consuming. Leveraging unsupervised learning, active learning, and synthetic data—alongside domain adaptation and knowledge distillation—can reduce annotation burdens while maintaining quality.

\vspace{3px}
\noindent\textbf{Model Development:} Five key areas require focused research to improve model performance and reliability.
(1) Architecture: Emerging models such as Transformers and Mamba offer promising capabilities. Their adaptation to aerial surveillance tasks should be systematically explored.
(2) Trustworthiness: Aerial data is often noisy and degraded. Models must be resilient to blur, low light, and weather effects, which are common in real-world scenarios.
(3) Robustness: Surveillance models are vulnerable to adversarial attacks. Understanding and mitigating these threats is essential for secure deployment.
(4) Uncertainty: Predictive confidence is crucial, especially in high-risk applications. Research should focus on quantifying and interpreting model uncertainty.
(5) Explainability: Most models operate as black boxes. Explainable AI (XAI) techniques can enhance transparency, enabling users to understand and trust model decisions.

\vspace{3px}
\noindent\textbf{Deployment-aware Model Development:} Practical deployment demands models that are both efficient and adaptive.
(1) Lightweight Models: Techniques like pruning, quantization, and distillation help reduce computational load, making models suitable for edge devices and UAVs.
(2) Onboard Processing: Real-time inference on resource-constrained platforms is critical. Designing models for onboard execution can reduce latency, power consumption, and reliance on ground stations.

\vspace{3px}
\noindent\textbf{Privacy and Ethics:} 
While aerial surveillance offers powerful capabilities, it also raises pressing ethical concerns. Real-world deployments - such as drone monitoring of protests or border regions - have sparked debates around civil liberties and informed consent. Moreover, algorithmic bias in biometric recognition may disproportionately affect certain demographic groups, leading to unfair outcomes. 
To ensure responsible development and deployment, we highlight four key priorities: 
\begin{itemize}
    \item \textit{(1) {Privacy Protection:}} Techniques such as federated learning, differential privacy, and onboard anonymization (e.g., face blurring) can safeguard personal data during collection and processing. 
    \item \textit{(2) {Ethical Design:}} Systems should incorporate explainable AI (XAI), fairness audits, and uncertainty estimation to promote transparency and accountability, especially in high-stakes scenarios. 
    \item \textit{(3) {Regulatory Compliance:}} Aerial systems must align with legal frameworks, including geofencing, audit trails, and local privacy laws, to prevent misuse and ensure lawful operation. 
    \item \textit{(4) {Stakeholder Engagement:}} Involving ethicists, policymakers, and affected communities is essential to ensure that surveillance technologies respect societal values and civil liberties. These measures aim to balance technological advancement with ethical responsibility, fostering public trust and minimizing harm.
\end{itemize}

\section{CONCLUDING REMARKS}
\label{sec:Conclusion}
Human-centric aerial surveillance offers notable advantages in scale, mobility, and adaptability, enabling effective monitoring in challenging environments. Recent datasets and initiatives like VisDrones and BRIAR have spurred academic interest, yet the field faces persistent challenges due to imaging limitations such as low resolution, extreme angles, and environmental noise. This survey provides the first comprehensive analysis of three aerial surveillance core tasks—detection, identification, and re-identification—within a structured framework of challenges, datasets, approaches, and techniques, highlighting current progress and open issues. Despite ongoing challenges, the outlook is promising. By consolidating a decade of research, this work aims to guide future efforts toward more accurate, resilient, and ethically grounded aerial surveillance solutions.

{
\bibliographystyle{ieee}
\bibliography{ASAS}

\begin{thebibliography}{100}\itemsep=-1pt

\bibitem{electronics11071151}
K.~R. Akshatha, A.~K. Karunakar, S.~B. Shenoy, A.~K. Pai, N.~H. Nagaraj, and S.~S. Rohatgi.
\newblock Human detection in aerial thermal images using faster r-cnn and ssd algorithms.
\newblock {\em Electronics}, 11(7), 2022.

\bibitem{EfficientFaceV2S}
M.~Alansari, K.~Alnuaimi, I.~Ganapathi, S.~Alansari, S.~Javed, A.~Shoufan, Y.~Zweiri, and N.~Werghi.
\newblock Efficientfacev2s: A lightweight model and a benchmarking approach for drone-captured face recognition.
\newblock {\em Expert Systems with Applications}, 273:126786, 2025.

\bibitem{SyNet}
B.~M. Albaba and S.~Ozer.
\newblock Synet: An ensemble network for object detection in {UAV} images.
\newblock In {\em ICPR}, 2020.

\bibitem{PersonReIDUAVsurvey}
Y.~Albaluchi, B.~Fu, N.~Damer, R.~Ramachandra, and K.~Raja.
\newblock Uav-based person re-identification: A survey of uav datasets, approaches, and challenges.
\newblock {\em Computer Vision and Image Understanding}, 251:104261, 2025.

\bibitem{MRFR}
G.~Amato, F.~Falchi, C.~Gennaro, F.~V. Massoli, and C.~Vairo.
\newblock Multi-resolution face recognition with drones.
\newblock In {\em International Conference on Sensors, Signal and Image Processing}, 2020.

\bibitem{HumanDetectionUAVsurvey}
A.~A. Bany~Abdelnabi and G.~Rabadi.
\newblock Human detection from unmanned aerial vehicles’ images for search and rescue missions: A state-of-the-art review.
\newblock {\em IEEE Access}, 12:152009--152035, 2024.

\bibitem{bolme2024data}
D.~S. Bolme, D.~Aykac, R.~Shivers, J.~Brogan, N.~Barber, B.~Zhang, L.~Davies, and D.~Cornett.
\newblock From data to insights: A covariate analysis of the iarpa briar dataset for multimodal biometric recognition algorithms at altitude and range.
\newblock In {\em IJCB}, 2024.

\bibitem{BIRDSAI}
E.~Bondi, R.~Jain, P.~Aggrawal, et~al.
\newblock Birdsai: A dataset for detection and tracking in aerial thermal infrared videos.
\newblock In {\em WACV}, 2020.

\bibitem{AUAIR}
I.~{Bozcan} and E.~{Kayacan}.
\newblock Au-air: A multi-modal {UAV} dataset for low altitude traffic surveillance.
\newblock In {\em ICRA}, 2020.

\bibitem{HERIDAL}
D.~Bozic-Stulic, Z.~Marusic, and S.~Gotovac.
\newblock Deep learning approach in aerial imagery for supporting land search and rescue missions.
\newblock {\em International Journal on Computer Vision}, 127(9):1256--1278, 2019.

\bibitem{CascadeRCNN}
Z.~{Cai} and N.~{Vasconcelos}.
\newblock {Cascade R-CNN: Delving Into High Quality Object Detection}.
\newblock In {\em CVPR}, 2018.

\bibitem{VisDroneDET2021}
Y.~Cao, Z.~He, and et~al.
\newblock Visdrone-det2021: The vision meets drone object detection challenge results.
\newblock In {\em ICCV Workshop}, pages 2847--2854, 2021.

\bibitem{RRNet}
C.~{Chen}, Y.~{Zhang}, Q.~{Lv}, S.~{Wei}, X.~{Wang}, X.~{Sun}, and J.~{Dong}.
\newblock Rrnet: A hybrid detector for object detection in drone-captured images.
\newblock In {\em ICCV Workshop}, 2019.

\bibitem{RotTrans}
S.~Chen, M.~Ye, and B.~Du.
\newblock Rotation invariant transformer for recognizing object in uavs.
\newblock In {\em ACM MM}, pages 2565--2574, 2022.

\bibitem{GANdetection}
Y.~Chen, J.~Li, Y.~Niu, and J.~He.
\newblock Small object detection networks based on classification-oriented super-resolution gan for {UAV} aerial imagery.
\newblock In {\em Chinese Control And Decision Conference}, 2019.

\bibitem{BRIARdataset}
D.~Cornett, J.~Brogan, and et~al.
\newblock Expanding accurate person recognition to new altitudes and ranges: The briar dataset.
\newblock In {\em WACV Workshop}, pages 593--602, 2023.

\bibitem{DBFR}
A.~Deeb, K.~Roy, and K.~D. Edoh.
\newblock Drone-based face recognition using deep learning.
\newblock In {\em Advanced Machine Learning Technologies and Applications}, 2021.

\bibitem{ArcFace}
J.~{Deng}, J.~{Guo}, N.~{Xue}, and S.~{Zafeiriou}.
\newblock Arcface: Additive angular margin loss for deep face recognition.
\newblock In {\em CVPR}, 2019.

\bibitem{GLSA}
S.~{Deng}, S.~{Li}, K.~{Xie}, W.~{Song}, X.~{Liao}, A.~{Hao}, and H.~{Qin}.
\newblock A global-local self-adaptive network for drone-view object detection.
\newblock {\em IEEE Transactions on Image Processing}, 30:1556--1569, 2021.

\bibitem{SODA}
K.~{Ding}, G.~{He}, H.~{Gu}, Z.~{Zhong}, S.~{Xiang}, and C.~{Pan}.
\newblock Train in dense and test in sparse: A method for sparse object detection in aerial images.
\newblock {\em IEEE Geoscience and Remote Sensing Letters}, 2020.

\bibitem{UAV-Gait}
T.~Ding, Q.~Zhao, F.~Liu, H.~Zhang, and P.~Peng.
\newblock A dataset and method for gait recognition with unmanned aerial vehicless.
\newblock In {\em ICME}, pages 1--6, 2022.

\bibitem{SRFR}
M.~Dosi, U.~Rathore, C.~Chiranjeev, A.~Agarwal, R.~Singh, and M.~Vatsa.
\newblock Is face super resolution truly pushing the boundaries of face recognition?
\newblock In {\em 2024 IEEE International Joint Conference on Biometrics (IJCB)}, pages 1--9, 2024.

\bibitem{FSAF}
B.~Du, Y.~Huang, J.~Chen, and D.~Huang.
\newblock Adaptive sparse convolutional networks with global context enhancement for faster object detection on drone images.
\newblock In {\em CVPR}, pages 13435--13444, 2023.

\bibitem{VisDroneDET2020}
D.~Du et~al.
\newblock {VisDrone-DET2020: The Vision Meets Drone Object Detection in Image Challenge Results}.
\newblock In {\em ECCV Workshop}, 2020.

\bibitem{CenterNet}
K.~{Duan}, S.~{Bai}, L.~{Xie}, H.~{Qi}, Q.~{Huang}, and Q.~{Tian}.
\newblock Centernet: Keypoint triplets for object detection.
\newblock In {\em ICCV}, 2019.

\bibitem{E2EunconstrainedFR}
J.~A. {Duncan}, N.~D. {Kalka}, B.~{Maze}, and A.~K. {Jain}.
\newblock End-to-end protocols and performance metrics for unconstrained face recognition.
\newblock In {\em ICB}, 2019.

\bibitem{MAVREC}
A.~Dutta, S.~Das, J.~Nielsen, R.~Chakraborty, and M.~Shah.
\newblock Multiview aerial visual recognition (mavrec): Can multi-view improve aerial visual perception?
\newblock In {\em CVPR}, pages 22678--22690, 2024.

\bibitem{Res2Net}
S.~H. {Gao}, M.~M. {Cheng}, K.~{Zhao}, X.~Y. {Zhang}, M.~H. {Yang}, and P.~{Torr}.
\newblock Res2net: A new multi-scale backbone architecture.
\newblock {\em T-PAMI}, 43(2):652--662, 2021.

\bibitem{MARN}
S.~{Gong}, Y.~{Shi}, and A.~{Jain}.
\newblock {Low Quality Video Face Recognition: Multi-Mode Aggregation Recurrent Network (MARN)}.
\newblock In {\em ICCV Workshop}, 2019.

\bibitem{REAN}
S.~Gong, Y.~Shi, A.~K. Jain, and N.~D. Kalka.
\newblock Recurrent embedding aggregation network for video face recognition.
\newblock {\em arXiv}, 1904.12019, 2019.

\bibitem{CFAN}
S.~{Gong}, Y.~{Shi}, N.~D. {Kalka}, and A.~K. {Jain}.
\newblock {Video Face Recognition: Component-wise Feature Aggregation Network (C-FAN)}.
\newblock In {\em ICB}, 2019.

\bibitem{FusionFactor}
Y.~Gong, X.~Yu, Y.~Ding, X.~Peng, J.~Zhao, and Z.~Han.
\newblock Effective fusion factor in fpn for tiny object detection.
\newblock In {\em WACV}, 2021.

\bibitem{AdaptiveFeat}
Z.~{Gong} and D.~{Li}.
\newblock {Towards Better Object Detection in Scale Variation with Adaptive Feature Selection}.
\newblock {\em arXiv}, 2012.03265, 2020.

\bibitem{DroneHIT}
A.~Grigorev, Z.~Tian, S.~Rho, J.~Xiong, S.~Liu, and F.~Jiang.
\newblock Deep person re-identification in {UAV} images.
\newblock {\em EURASIP Journal on Advances in Signal Processing}, 2019(1):54 -- 64, 2019.

\bibitem{CRFR}
K.~Grm, B.~Kemal~Özata, A.~Kantarci, V.~Štruc, and H.~Kemal~Ekenel.
\newblock Degrade or super-resolve to recognize? bridging the domain gap for cross-resolution face recognition.
\newblock {\em IEEE Access}, 13:10542--10558, 2025.

\bibitem{LidarReID}
W.~Guo, Z.~Pan, Y.~Liang, Z.~Xi, Z.~Zhong, J.~Feng, and J.~Zhou.
\newblock Lidar-based person re-identification.
\newblock In {\em CVPR}, pages 17437--17447, 2024.

\bibitem{DetectionThermalFCRN}
A.~Haider, F.~Shaukat, and J.~Mir.
\newblock Human detection in aerial thermal imaging using a fully convolutional regression network.
\newblock {\em Infrared Physics and Technology}, 116, 2021.

\bibitem{VisIRreID2}
Y.~Hao, N.~Wang, J.~Li, and X.~Gao.
\newblock Hsme: Hypersphere manifold embedding for visible thermal person re-identification.
\newblock In {\em AAAI}, volume~33, 2019.

\bibitem{ALSS-YOLO}
A.~He, X.~Li, X.~Wu, C.~Su, J.~Chen, S.~Xu, and X.~Guo.
\newblock Alss-yolo: An adaptive lightweight channel split and shuffling network for tir wildlife detection in uav imagery.
\newblock {\em IEEE Journal of Selected Topics in Applied Earth Observations and Remote Sensing}, 17:17308--17326, 2024.

\bibitem{ResNet}
K.~He, X.~Zhang, S.~Ren, and J.~Sun.
\newblock Deep residual learning for image recognition.
\newblock In {\em CVPR}, 2016.

\bibitem{PLAOD}
S.~{Hong}, S.~{Kang}, and D.~{Cho}.
\newblock Patch-level augmentation for object detection in aerial images.
\newblock In {\em ICCV Workshop}, 2019.

\bibitem{DroneFace}
H.-J. Hsu and K.-T. Chen.
\newblock {DroneFace: An Open Dataset for Drone Research}.
\newblock In {\em ACM on Multimedia Systems Conference}, 2017.

\bibitem{LATex}
X.~Hu, Y.~Wang, P.~Zhang, and H.~Lu.
\newblock Latex: Leveraging attribute-based text knowledge for aerial-ground person re-identification, 2025.

\bibitem{SDReID}
X.~Hu, P.~Zhang, Y.~Wang, B.~Yan, and H.~Lu.
\newblock Sd-reid: View-aware stable diffusion for aerial-ground person re-identification, 2025.

\bibitem{MRReID}
M.~Huang, C.~Hou, X.~Zheng, and Z.~Wang.
\newblock Multi-resolution feature perception network for uav person re-identification.
\newblock {\em Multimedia Tools and Applications}, 2024.

\bibitem{BRIARNet}
S.~Huang, R.~P. Kathirvel, Y.~Guo, C.~P. Lau, and R.~Chellappa.
\newblock Whole-body detection, identification and recognition at altitude and range.
\newblock {\em IEEE Transactions on Biometrics, Behavior, and Identity Science}, 7(3):331--343, 2025.

\bibitem{BRIAR}
I.~A. R. P.~A. (IARPA).
\newblock {Biometric Recognition and Identification at Altitude and Range (BRIAR)}, 2021.

\bibitem{CoNAN}
B.~Jawade, D.~Dayal~Mohan, P.~Shetty, D.~Fedorishin, S.~Setlur, and V.~Govindaraju.
\newblock Conan: Conditional neural aggregation network for unconstrained long range biometric feature fusion.
\newblock {\em IEEE Transactions on Biometrics, Behavior, and Identity Science}, 6(4):602--612, 2024.

\bibitem{ProxyFusion}
B.~Jawade, A.~Stone, D.~D. Mohan, X.~Wang, S.~Setlur, and V.~Govindaraju.
\newblock Proxyfusion: face feature aggregation through sparse experts.
\newblock In {\em NIPS}, 2025.

\bibitem{MFFSODNet}
L.~Jiang, B.~Yuan, J.~Du, B.~Chen, H.~Xie, J.~Tian, and Z.~Yuan.
\newblock Mffsodnet: Multiscale feature fusion small object detection network for uav aerial images.
\newblock {\em IEEE Transactions on Instrumentation and Measurement}, 73:1--14, 2024.

\bibitem{AdaptiveAnchor}
R.~{Jin} and D.~{Lin}.
\newblock Adaptive anchor for fast object detection in aerial image.
\newblock {\em IEEE Geoscience and Remote Sensing Letters}, 17(5):839--843, 2020.

\bibitem{10423161}
I.~Joshi, M.~Grimmer, C.~Rathgeb, C.~Busch, F.~Bremond, and A.~Dantcheva.
\newblock Synthetic data in human analysis: A survey.
\newblock {\em T-PAMI}, 46(7):4957--4976, 2024.

\bibitem{IJB-S}
N.~D. {Kalka}, B.~{Maze}, J.~A. {Duncan}, K.~{O’Connor}, S.~{Elliott}, K.~{Hebert}, J.~{Bryan}, and A.~K. {Jain}.
\newblock Ijb–s: Iarpa janus surveillance video benchmark.
\newblock In {\em BTAS}, 2018.

\bibitem{DroneSURF}
I.~{Kalra}, M.~{Singh}, S.~{Nagpal}, R.~{Singh}, M.~{Vatsa}, and P.~B. {Sujit}.
\newblock Dronesurf: Benchmark dataset for drone-based face recognition.
\newblock In {\em FG}, pages 1--7, 2019.

\bibitem{MaCVi}
B.~Kiefer, M.~Kristan, and et~al.
\newblock 1st workshop on maritime computer vision (macvi) 2023: Challenge results.
\newblock In {\em WACV Workshop}, pages 265--302, 2023.

\bibitem{DomainLabels}
B.~Kiefer, M.~Messmer, and A.~Zell.
\newblock Leveraging domain labels for object detection from {UAVs}.
\newblock {\em arXiv}, 2101.12677, 2021.

\bibitem{MaCVi2}
B.~Kiefer, L.~\v{Z}ust, and et~al.
\newblock 2nd workshop on maritime computer vision (macvi) 2024: Challenge results.
\newblock In {\em WACV Workshops}, pages 869--891, January 2024.

\bibitem{PDESTRE}
S.~V.~A. Kumar, E.~Yaghoubi, A.~Das, B.~S. Harish, and H.~Proença.
\newblock The p-destre: A fully annotated dataset for pedestrian detection, tracking, and short/long-term re-identification from aerial devices.
\newblock {\em IEEE Transactions on Information Forensics and Security}, 2021.

\bibitem{ClusterNet}
R.~{LaLonde}, D.~{Zhang}, and M.~{Shah}.
\newblock Clusternet: Detecting small objects in large scenes by exploiting spatio-temporal information.
\newblock In {\em CVPR}, 2018.

\bibitem{CornerNet}
H.~Law and J.~Deng.
\newblock Cornernet: detecting objects as paired keypoints.
\newblock In {\em ECCV}, 2018.

\bibitem{MRP}
R.~Layne, T.~Hospedales, and S.~Gong.
\newblock Investigating open-world person re-identification using a drone.
\newblock In {\em ECCVWorkshop}, 2015.

\bibitem{lee2024exploringimpactsyntheticdata}
H.~Lee, Y.~Zhang, Y.-T. Shen, H.~Kwon, and S.~S. Bhattacharyya.
\newblock Exploring the impact of synthetic data for aerial-view human detection, 2024.

\bibitem{AgriDrone}
A.~Leipnitz, T.~Strutz, and O.~Jokisch.
\newblock Spatial resolution-independent{CNN}-based person detection in agricultural image data.
\newblock In {\em Interactive Collaborative Robotics}, pages 189--199, 2020.

\bibitem{DroneGait}
A.~Li, S.~Hou, Q.~Cai, Y.~Fu, and Y.~Huang.
\newblock Gait recognition with drones: A benchmark.
\newblock {\em IEEE Transactions on Multimedia}, 26:3530--3540, 2024.

\bibitem{AerialGait}
A.~Li, S.~Hou, C.~Wang, Q.~Cai, and Y.~Huang.
\newblock Aerialgait: Bridging aerial and ground views for gait recognition.
\newblock In {\em ACM International Conference on Multimedia}, page 1139–1147, 2024.

\bibitem{DMNet}
C.~Li, T.~Yang, S.~Zhu, C.~Chen, and S.~Guan.
\newblock Density map guided object detection in aerial images.
\newblock In {\em CVPR Workshop}, 2020.

\bibitem{UAV-Human}
T.~Li, J.~Liu, W.~Zhang, Y.~Ni, W.~Wang, and Z.~Li.
\newblock {UAV-Human: A Large Benchmark for Human Behavior Understanding with UAVs}.
\newblock In {\em CVPR}, 2021.

\bibitem{CVVNet}
X.~Li, W.~Song, Y.~Huang, W.~Meng, and L.~Chang.
\newblock Cvvnet: A cross-vertical-view network for gait recognition.
\newblock {\em arXiv}, 2025.

\bibitem{ContextAerial}
X.~{Liang}, J.~{Zhang}, L.~{Zhuo}, Y.~{Li}, and Q.~{Tian}.
\newblock Small object detection in {UAV} images using feature fusion and scaling-based single shot detector with spatial context analysis.
\newblock {\em IEEE Transactions on Circuits and Systems for Video Technology}, 30(6):1758--1770, 2020.

\bibitem{USFA}
H.~Lin, J.~Zhou, Y.~Gan, C.-M. Vong, and Q.~Liu.
\newblock Novel up-scale feature aggregation for object detection in aerial images.
\newblock {\em Neurocomputing}, 411:364--374, 2020.

\bibitem{FPN}
T.~{Lin}, P.~{Dollár}, R.~{Girshick}, K.~{He}, B.~{Hariharan}, and S.~{Belongie}.
\newblock Feature pyramid networks for object detection.
\newblock In {\em CVPR}, 2017.

\bibitem{RetinaNet}
T.-Y. Lin, P.~Goyal, R.~Girshick, K.~He, and P.~Dollar.
\newblock Focal loss for dense object detection.
\newblock In {\em ICCV}, 2017.

\bibitem{FarSight}
F.~Liu, R.~Ashbaugh, and et~al.
\newblock Farsight: A physics-driven whole-body biometric system at large distance and altitude.
\newblock In {\em WACV}, 2024.

\bibitem{FarSight2}
F.~Liu, N.~Chimitt, L.~Guo, J.~Jain, A.~Kane, M.~Kim, W.~Robbins, Y.~Su, D.~Ye, X.~Zhang, J.~Zhu, S.~Satyakam, C.~Perry, S.~H. Chan, A.~Ross, H.~Shi, Z.~Wang, A.~Jain, and X.~Liu.
\newblock Person recognition at altitude and range: Fusion of face, body shape and gait.
\newblock {\em arXiv}, 2025.

\bibitem{SmallDetSurvey2}
Y.~Liu, P.~Sun, N.~Wergeles, and Y.~Shang.
\newblock A survey and performance evaluation of deep learning methods for small object detection.
\newblock {\em Expert Systems with Applications}, 172:114602, 2021.

\bibitem{BoT}
H.~Luo, Y.~Gu, X.~Liao, S.~Lai, and W.~Jiang.
\newblock {Bag of Tricks and a Strong Baseline for Deep Person Re-Identification}.
\newblock In {\em CVPR Workshop}, 2019.

\bibitem{HERIDAL2}
Z.~{Marusic}, D.~{Bozic-Stulic}, S.~{Gotovac}, and T.~{Marusic}.
\newblock Region proposal approach for human detection on aerial imagery.
\newblock In {\em International Conference on Smart and Sustainable Technologies}, 2018.

\bibitem{marvasti2020comet}
S.~M. Marvasti-Zadeh, J.~Khaghani, H.~Ghanei-Yakhdan, S.~Kasaei, and L.~Cheng.
\newblock {COMET}: context-aware {IoU}-guided network for small object tracking.
\newblock In {\em ACCV}, 2020.

\bibitem{ScaleInvarianceAerial}
M.~Messmer, B.~Kiefer, and A.~Zell.
\newblock Gaining scale invariance in {UAV} bird's eye view object detection by adaptive resizing.
\newblock {\em arXiv}, 2101.12694, 2021.

\bibitem{EyesInTheSky}
A.~H. Michel.
\newblock {\em Eyes in the Sky}.
\newblock Mariner Books, 2019.

\bibitem{ReIDdesign}
L.~Moritz, A.~Specker, and A.~Schumann.
\newblock {A study of person re-identification design characteristics for aerial data}.
\newblock In {\em Pattern Recognition and Tracking XXXII}, 2021.

\bibitem{ContextRole}
R.~{Mottaghi}, X.~{Chen}, X.~{Liu}, N.~{Cho}, S.~{Lee}, S.~{Fidler}, R.~{Urtasun}, and A.~{Yuille}.
\newblock The role of context for object detection and semantic segmentation in the wild.
\newblock In {\em CVPR}, 2014.

\bibitem{LRface}
S.~P. Mudunuri and S.~Biswas.
\newblock Low resolution face recognition across variations in pose and illumination.
\newblock {\em T-PAMI}, 38(5):1034--1040, 2016.

\bibitem{UAV123}
M.~Mueller, N.~Smith, and B.~Ghanem.
\newblock A benchmark and simulator for {UAV} tracking.
\newblock In {\em ECCV}, pages 445--461, 2016.

\bibitem{Hourglass}
A.~Newell, K.~Yang, and J.~Deng.
\newblock Stacked hourglass networks for human pose estimation.
\newblock In {\em ECCV}, 2016.

\bibitem{AGVPReID2025}
H.~Nguyen, K.~Nguyen, L.~Feng, P.~Akila, C.~Fookes, and S.~Sridharan.
\newblock Ag-vpreid: A challenging large-scale benchmark for aerial-ground video-based person re-identification.
\newblock In {\em CVPR}, 2025.

\bibitem{AGReID2023}
H.~Nguyen, K.~Nguyen, C.~Fookes, and S.~Sridharan.
\newblock Aerial-ground person re-id.
\newblock In {\em ICME}, 2023.

\bibitem{AGReID_Challenge}
H.~Nguyen, K.~Nguyen, C.~Fookes, and S.~Sridharan.
\newblock Ag-reid 2023: Aerial-ground person re-identification challenge results.
\newblock In {\em IJCB}, 2023.

\bibitem{AGReIDv2}
H.~Nguyen, K.~Nguyen, S.~Sridharan, and C.~Fookes.
\newblock Ag-reid.v2: Bridging aerial and ground views for person re-identification.
\newblock {\em IEEE Transactions on Information Forensics and Security}, 19:2896--2908, 2024.

\bibitem{FirstAerialPersonReID}
O.~{Oreifej}, R.~{Mehran}, and M.~{Shah}.
\newblock Human identity recognition in aerial images.
\newblock In {\em CVPR}, 2010.

\bibitem{AerialFace}
Z.~Ou, L.~Yao, T.~Wu, and F.~Liu.
\newblock Aerialface: A light weight framework for unmanned aerial vehicle face recognition.
\newblock In {\em FG}, pages 1--7, 2024.

\bibitem{LibraRCNN}
J.~{Pang}, K.~{Chen}, J.~{Shi}, H.~{Feng}, W.~{Ouyang}, and D.~{Lin}.
\newblock {Libra R-CNN: Towards Balanced Learning for Object Detection}.
\newblock In {\em CVPR}, 2019.

\bibitem{VGGFace}
O.~M. Parkhi, A.~Vedaldi, and A.~Zisserman.
\newblock Deep face recognition.
\newblock In {\em BMVC}, 2015.

\bibitem{perera2018human}
A.~G. Perera, Y.~W. Law, and J.~Chahl.
\newblock Human pose and path estimation from aerial video using dynamic classifier selection.
\newblock {\em Cognitive Computation}, 10(6):1019--1041, 2018.

\bibitem{SelectiveTiling}
G.~Plastiras, C.~Kyrkou, and T.~Theocharides.
\newblock Efficient convnet-based object detection for {UAVs} by selective tile processing.
\newblock In {\em International Conference on Distributed Smart Cameras}, 2019.

\bibitem{Qiu2024}
J.~Qiu, Z.~Feng, L.~Wang, and J.~Lai.
\newblock Salient part-aligned and keypoint disentangling transformer for person re-identification in aerial imagery.
\newblock In {\em ICME}, 2024.

\bibitem{FasterRCNN}
S.~Ren, K.~He, R.~Girshick, and J.~Sun.
\newblock {Faster R-CNN: Towards real-time object detection with region proposal networks}.
\newblock In {\em NeuRIPS}, 2015.

\bibitem{StanfordDrones}
A.~Robicquet, A.~Sadeghian, A.~Alahi, and S.~Savarese.
\newblock Learning social etiquette: Human trajectory understanding in crowded scenes.
\newblock In {\em ECCV}, 2016.

\bibitem{SARD}
S.~{Sambolek} and M.~{Ivasic-Kos}.
\newblock Automatic person detection in search and rescue operations using deep {CNN} detectors.
\newblock {\em IEEE Access}, 9:37905--37922, 2021.

\bibitem{SearchRescureThermal}
D.~C. Schedl, I.~Kurmi, and O.~Bimber.
\newblock Search and rescue with airborne optical sectioning.
\newblock {\em Nature Machine Intelligence}, 2020.

\bibitem{FaceNet}
F.~{Schroff}, D.~{Kalenichenko}, and J.~{Philbin}.
\newblock Facenet: A unified embedding for face recognition and clustering.
\newblock In {\em CVPR}, 2015.

\bibitem{AerialPersonReID_CovFu}
A.~Schumann and J.~Metzler.
\newblock {Person re-identification across aerial and ground-based cameras by deep feature fusion}.
\newblock In {\em Automatic Target Recognition XXVII}, 2017.

\bibitem{AerialPersonReID_Attri}
A.~Schumann and J.~Metzler.
\newblock {Adapted deep feature fusion for person re-identification in aerial images}.
\newblock In {\em Autonomous Systems: Sensors, Vehicles, Security, and the Internet of Everything}, 2018.

\bibitem{sien2019deep}
J.~P.~T. Sien, K.~H. Lim, and P.-I. Au.
\newblock Deep learning in gait recognition for drone surveillance system.
\newblock In {\em Materials Science and Engineering}, 2019.

\bibitem{AVI}
A.~Singh, D.~Patil, and S.~Omkar.
\newblock Eye in the sky: Real-time drone surveillance system for violent individuals identification using scatternet hybrid deep learning.
\newblock In {\em CVPR Workshop}, 2018.

\bibitem{DeriveNet}
M.~Singh, S.~Nagpal, R.~Singh, and M.~Vatsa.
\newblock Derivenet for (very) low resolution image classification.
\newblock {\em T-PAMI}, 44(10):6569--6577, 2022.

\bibitem{COSAM}
A.~{Subramaniam}, A.~{Nambiar}, and A.~{Mittal}.
\newblock Co-segmentation inspired attention networks for video-based person re-identification.
\newblock In {\em ICCV}, 2019.

\bibitem{InceptionNet}
C.~{Szegedy}, {Wei Liu}, {Yangqing Jia}, P.~{Sermanet}, S.~{Reed}, D.~{Anguelov}, D.~{Erhan}, V.~{Vanhoucke}, and A.~{Rabinovich}.
\newblock Going deeper with convolutions.
\newblock In {\em CVPR}, 2015.

\bibitem{EfficientDet}
M.~Tan, R.~Pang, and Q.~Le.
\newblock {EfficientDet}: Scalable and efficient object detection.
\newblock In {\em CVPR}, 2020.

\bibitem{PENet}
Z.~{Tang}, X.~{Liu}, and B.~{Yang}.
\newblock Penet: Object detection using points estimation in high definition aerial images.
\newblock In {\em ICMLA}, 2020.

\bibitem{FCOS}
Z.~Tian, C.~Shen, H.~Chen, and T.~He.
\newblock {FCOS}: Fully convolutional one-stage object detection.
\newblock In {\em ICCV}, 2019.

\bibitem{FCOSPAMI}
Z.~Tian, C.~Shen, H.~Chen, and T.~He.
\newblock {FCOS}: A simple and strong anchor-free object detector.
\newblock {\em T-PAMI}, 2021.

\bibitem{SmallDetSurvey}
K.~Tong, Y.~Wu, and F.~Zhou.
\newblock Recent advances in small object detection based on deep learning: A review.
\newblock {\em Image and Vision Computing}, 97:103910, 2020.

\bibitem{MaCVi3}
E.~K.~U. Tübingen.
\newblock 3rd workshop on maritime computer vision (macvi).
\newblock {https://macvi.org/workshop/macvi25}, 2025.

\bibitem{TilingPower}
F.~O. {Unel}, B.~O. {Ozkalayci}, and C.~{Cigla}.
\newblock The power of tiling for small object detection.
\newblock In {\em CVPR Workshop}, 2019.

\bibitem{Seadronessee}
L.~A. Varga, B.~Kiefer, M.~Messmer, and A.~Zell.
\newblock Seadronessee: A maritime benchmark for detecting humans in open water.
\newblock In {\em WACV}, pages 2260--2270, 2022.

\bibitem{LGMloss}
W.~Wan, Y.~Zhong, T.~Li, and J.~Chen.
\newblock Rethinking feature distribution for loss functions in image classification.
\newblock In {\em CVPR}, 2018.

\bibitem{SAMR}
H.~{Wang}, Z.~{Wang}, M.~{Jia}, A.~{Li}, T.~{Feng}, W.~{Zhang}, and L.~{Jiao}.
\newblock Spatial attention for multi-scale feature refinement for object detection.
\newblock In {\em ICCV Workshop}, 2019.

\bibitem{HRNet}
J.~Wang, K.~Sun, T.~Cheng, B.~Jiang, C.~Deng, Y.~Zhao, D.~Liu, Y.~Mu, M.~Tan, X.~Wang, W.~Liu, and B.~Xiao.
\newblock Deep high-resolution representation learning for visual recognition.
\newblock {\em T-PAMI}, 2019.

\bibitem{DG-UAVOD}
K.~Wang, X.~Fu, Y.~Huang, C.~Cao, G.~Shi, and Z.-J. Zha.
\newblock Generalized uav object detection via frequency domain disentanglement.
\newblock In {\em CVPR}, pages 1064--1073, 2023.

\bibitem{Pelee}
R.~J. Wang, X.~Li, and C.~X. Ling.
\newblock Pelee: A real-time object detection system on mobile devices.
\newblock In {\em NeuRIPS}, 2018.

\bibitem{SeCap}
S.~Wang, Y.~Wang, R.~Wu, B.~Jiao, W.~Wang, and P.~Wang.
\newblock Secap: Self-calibrating and adaptive prompts for cross-view person re-identification in aerial-ground networks.
\newblock In {\em CVPR}, 2025.

\bibitem{DTST}
Y.~Wang and M.~Pishgar.
\newblock Dynamic token selection for aerial-ground person re-identification.
\newblock In {\em ICME}, 2025.

\bibitem{AdaptiveSearching}
Y.~Wang, Y.~Yang, and X.~Zhao.
\newblock Object detection using clustering algorithm adaptive searching regions in aerial images.
\newblock In {\em ECCVWorkshop}, 2020.

\bibitem{VisIRreID3}
Z.~{Wang}, Z.~{Wang}, Y.~{Zheng}, Y.~{Chuang}, and S.~{Satoh}.
\newblock Learning to reduce dual-level discrepancy for infrared-visible person re-identification.
\newblock In {\em CVPR}, 2019.

\bibitem{SPSNet}
H.~Wei, Q.~Li, J.~Pan, J.~Chen, Y.~Zhang, L.~Qi, and Y.~Zhou.
\newblock Spsnet: semantic-guided perspective shift network for robust person re-identification in drone imagery: Spsnet: semantic-guided perspective shift...
\newblock {\em Vis. Comput.}, 41(8):5563–5582, Dec. 2024.

\bibitem{VisIRreID}
A.~{Wu}, W.~{Zheng}, H.~{Yu}, S.~{Gong}, and J.~{Lai}.
\newblock Rgb-infrared cross-modality person re-identification.
\newblock In {\em ICCV}, 2017.

\bibitem{NDFT}
Z.~{Wu}, K.~{Suresh}, P.~{Narayanan}, H.~{Xu}, H.~{Kwon}, and Z.~{Wang}.
\newblock Delving into robust object detection from {UAVs}: A deep nuisance disentanglement approach.
\newblock In {\em ICCV}, 2019.

\bibitem{MERI}
J.~Xu, W.~Wang, H.~Wang, and J.~Guo.
\newblock Multi-model ensemble with rich spatial information for object detection.
\newblock {\em Pattern Recognition}, 99:107098, 2020.

\bibitem{MetaTransfer}
L.~Xu, H.~Peng, L.~Wang, and D.~Xia.
\newblock Meta-transfer learning for person re-identification in aerial imagery.
\newblock In {\em Computer Supported Cooperative Work and Social Computing}, pages 634--644, 2023.

\bibitem{HumanAttriSurvey}
E.~Yaghoubi, F.~Khezeli, D.~Borza, S.~A. Kumar, J.~Neves, and H.~Proença.
\newblock Human attribute recognition - a comprehensive survey.
\newblock {\em Applied Sciences}, 10(16), 2020.

\bibitem{QueryDet}
C.~Yang, Z.~Huang, and N.~Wang.
\newblock Querydet: Cascaded sparse query for accelerating high-resolution small object detection.
\newblock {\em arXiv}, 2103.09136, 2021.

\bibitem{ClusterDet}
F.~Yang, H.~Fan, P.~Chu, E.~Blasch, and H.~Ling.
\newblock Clustered object detection in aerial images.
\newblock In {\em ICCV}, 2019.

\bibitem{ArbitraryOrientedDetection}
X.~Yang and J.~Yan.
\newblock Arbitrary-oriented object detection with circular smooth label.
\newblock {\em ECCV}, 2020.

\bibitem{ReIDsurvey}
M.~Ye, J.~Shen, G.~Lin, T.~Xiang, L.~Shao, and S.~C.~H. Hoi.
\newblock Deep learning for person re-identification: A survey and outlook.
\newblock {\em T-PAMI}, 2021.

\bibitem{UAVDT}
H.~Yu, G.~Li, W.~Zhang, Q.~Huang, D.~Du, Q.~Tian, and N.~Sebe.
\newblock The {UAV} benchmark: Object detection, tracking and baseline.
\newblock {\em International Journal of Computer Vision}, 128(5):1141 -- 1159, 2020.

\bibitem{DSHNet}
W.~Yu, T.~Yang, and C.~Chen.
\newblock Towards resolving the challenge of long-tail distribution in {UAV} images for object detection.
\newblock In {\em WACV}, 2021.

\bibitem{TinyPersonsChallenge}
X.~Yu et~al.
\newblock The 1st tiny object detection challenge: Methods and results.
\newblock In {\em ECCVWorkshop}, 2020.

\bibitem{SRface}
X.~Yu, B.~Fernando, R.~Hartley, and F.~Porikli.
\newblock Semantic face hallucination: Super-resolving very low-resolution face images with supplementary attributes.
\newblock {\em T-PAMI}, 42(11):2926--2943, 2020.

\bibitem{TinyPersons}
X.~{Yu}, Y.~{Gong}, N.~{Jiang}, Q.~{Ye}, and Z.~{Han}.
\newblock Scale match for tiny person detection.
\newblock In {\em WACV}, 2020.

\bibitem{FullyExploitAerial}
J.~{Zhang}, J.~{Huang}, X.~{Chen}, and D.~{Zhang}.
\newblock How to fully exploit the abilities of aerial image detectors.
\newblock In {\em ICCV Workshop}, 2019.

\bibitem{CARGO}
Q.~Zhang, L.~Wang, V.~M. Patel, X.~Xie, and J.~Lai.
\newblock View-decoupled transformer for person re-identification under aerial-ground camera network.
\newblock In {\em CVPR}, pages 22000--22009, June 2024.

\bibitem{GDFNet}
R.~Zhang, Z.~Shao, X.~Huang, J.~Wang, and D.~Li.
\newblock Object detection in {UAV} images via global density fused convolutional network.
\newblock {\em Remote Sensing}, 12(19), 2020.

\bibitem{ATSS}
S.~{Zhang}, C.~{Chi}, Y.~{Yao}, Z.~{Lei}, and S.~Z. {Li}.
\newblock Bridging the gap between anchor-based and anchor-free detection via adaptive training sample selection.
\newblock In {\em CVPR}, 2020.

\bibitem{G2AVReID}
S.~Zhang, W.~Luo, D.~Cheng, Q.~Yang, L.~Ran, Y.~Xing, and Y.~Zhang.
\newblock Cross-platform video person reid: A new benchmark dataset and adaptation approach.
\newblock In {\em ECCV}, 2024.

\bibitem{PRAI1581}
S.~Zhang, Q.~Zhang, Y.~Yang, X.~Wei, P.~Wang, B.~Jiao, and Y.~Zhang.
\newblock Person re-identification in aerial imagery.
\newblock {\em IEEE Transactions on Multimedia}, 2021.

\bibitem{DSOD}
X.~{Zhang}, E.~{Izquierdo}, and K.~{Chandramouli}.
\newblock Dense and small object detection in {UAV} vision based on cascade network.
\newblock In {\em ICCV Workshop}, 2019.

\bibitem{FreeAnchor}
X.~Zhang, F.~Wan, C.~Liu, R.~Ji, and Q.~Ye.
\newblock Freeanchor: Learning to match anchors for visual object detection.
\newblock In {\em NeuRIPS}, 2019.

\bibitem{MCNN}
Y.~{Zhang}, D.~{Zhou}, S.~{Chen}, S.~{Gao}, and Y.~{Ma}.
\newblock Single-image crowd counting via multi-column convolutional neural network.
\newblock In {\em CVPR}, 2016.

\bibitem{OSNet}
K.~Zhou, Y.~Yang, A.~Cavallaro, and T.~Xiang.
\newblock Omni-scale feature learning for person re-identification.
\newblock In {\em ICCV}, 2019.

\bibitem{VisDrones}
P.~Zhu, L.~Wen, D.~Du, X.~Bian, Q.~Hu, and H.~Ling.
\newblock Vision meets drones: Past, present and future.
\newblock {\em arXiv}, 2001.06303, 2021.

\bibitem{ObjectDetection20Years}
Z.~Zou, K.~Chen, Z.~Shi, Y.~Guo, and J.~Ye.
\newblock Object detection in 20 years: A survey.
\newblock {\em Proceedings of the IEEE}, 2023.

\end{thebibliography}
}

\end{document}